\documentclass{article}

\newif\ifieee
\ieeefalse

\usepackage{hyperref}
\hypersetup{
    colorlinks=true,
    linkcolor=blue,
    urlcolor=blue,
    citecolor=blue
}

\usepackage{url}

\usepackage[style=ieee,natbib=true]{biblatex}
\usepackage[margin=1.1in]{geometry}
\usepackage{authblk}                % \author
\usepackage{amsmath,amssymb,amsthm}
\usepackage{graphicx}
\usepackage{subfigure}
\usepackage{booktabs}

\usepackage{bbm}                    % \mathbb{1}
\usepackage{mathtools}              % \coloneqq
\usepackage{makecell}               % break cell in tables
\usepackage{xcolor}                 % allows using different text colors
\title{Pruning the Way to Reliable Policies:\\A Multi-Objective Deep Q-Learning Approach to Critical Care\footnote{This work has been published in the \textit{Journal of Biomedical and Health Informatics}.  
2168-2194 \textcopyright\ 2024 IEEE. Personal use is permitted, but republication/redistribution requires IEEE permission. \\
\textbf{Citation:} A. Shirali, A. Schubert and A. Alaa, ``Pruning the Way to Reliable Policies: A Multi-Objective Deep Q-Learning Approach to Critical Care,'' in IEEE Journal of Biomedical and Health Informatics, vol. 28, no. 10, pp. 6268-6279, Oct. 2024, doi: \url{https://doi.org/10.1109/JBHI.2024.3415115}. \\
\textbf{Accepted version available here:} \url{https://ieeexplore.ieee.org/document/10559219}}
}

\author[1]{Ali Shirali\thanks{Equal contribution}}
\author[1,2]{Alexander Schubert$^\dagger$}
\author[1,2]{Ahmed Alaa}
\affil[1]{University of California, Berkeley}
\affil[2]{University of California, San Francisco}

\date{}

\usepackage{amsmath,amsfonts,bm}

\DeclareMathOperator*{\softmax}{softmax}

\def\1{\bm{1}}

\def\vr{{\bm{r}}}

\def\vw{{\bm{w}}}

\DeclareMathAlphabet{\mathsfit}{\encodingdefault}{\sfdefault}{m}{sl}
\SetMathAlphabet{\mathsfit}{bold}{\encodingdefault}{\sfdefault}{bx}{n}

\def\gA{{\mathcal{A}}}
\def\gB{{\mathcal{B}}}

\def\gD{{\mathcal{D}}}

\def\gL{{\mathcal{L}}}

\def\gP{{\mathcal{P}}}

\def\gS{{\mathcal{S}}}

\def\gW{{\mathcal{W}}}

\def\sT{{\mathbb{T}}}

\newcommand{\E}{\mathbb{E}}

\newcommand{\R}{\mathbb{R}}

\DeclareMathOperator*{\argmax}{arg\,max}

\def\vQ{{\bm{Q}}}
\newcommand{\One}{\mathbbm{1}}

\newcommand{\sectionref}[1]{Section~\ref{#1}}
\newcommand{\equationref}[1]{Equation~\ref{#1}}
\newcommand{\figureref}[1]{Figure~\ref{#1}}
\newcommand{\tableref}[1]{Table~\ref{#1}}
\begin{document}

\maketitle

\begin{abstract}
Medical treatments often involve a sequence of decisions, each informed by previous outcomes. This process closely aligns with reinforcement learning (RL), a framework for optimizing sequential decisions to maximize cumulative rewards under unknown dynamics. While RL shows promise for creating data-driven treatment plans, its application in medical contexts is challenging due to the frequent need to use sparse rewards, primarily defined based on mortality outcomes. This sparsity can reduce the stability of offline estimates, posing a significant hurdle in fully utilizing RL for medical decision-making.  We introduce a deep Q-learning approach to obtain more reliable critical care policies by integrating relevant but noisy frequently measured biomarker signals into the reward specification without compromising the optimization of the main outcome. Our method prunes the action space based on all available rewards before training a final model on the sparse main reward. This approach minimizes potential distortions of the main objective while extracting valuable information from intermediate signals to guide learning. We evaluate our method in off-policy and offline settings using simulated environments and real health records from intensive care units. Our empirical results demonstrate that our method outperforms common offline RL methods such as conservative Q-learning and batch-constrained deep Q-learning. By disentangling sparse rewards and frequently measured reward proxies through action pruning, our work represents a step towards developing reliable policies that effectively harness the wealth of available information in data-intensive critical care environments.
\end{abstract}

\section{Introduction}

% Par 1: Motivate ICU setting and RL (in response to reviewer)
The intensive care unit (ICU) stands out as a prime candidate for the development of data-driven decision support tools. In the ICU, a comprehensive array of physiological patient biomarkers is continuously monitored and digitally stored, offering a rich set of information to guide decisions. Patients in the ICU typically experience critical health conditions, where accurate medical decisions are crucial and can be the determinant between life and death. Moreover, decision-making in this environment is inherently sequential, with a plethora of biomarkers offering valuable, albeit noisy, insights into the patient's evolving health status. Given these characteristics, there has been a surge of interest in exploring the prospects of reinforcement learning (RL) in critical care settings~\citep{henry2015targeted,prasad-2017,raghu2017continuous,raghu2018model,cheng2018optimal,komorowski2018artificial,lin-2018,peine-2021,mollura2022reinforcement,nanayakkara2022unifying, fatemi2021medical,tang2022leveraging,kamran2024evaluation,zhou2022personalized}, where the development of personalized, data-driven treatment policies for the management of vasopressors and intravenous fluids in sepsis patients has become a key area of interest~\citep{raghu2017continuous, komorowski2018artificial,peng-2018,fatemi2021medical,mollura2022reinforcement,nanayakkara2022unifying, liu2024value,wu2023value,yu2023towards}. 

% Par 2: Outline the challenge
Specifying an effective reward function is a central challenge in medical RL applications. While patient survival is a critical outcome of interest, a survival-based reward function is sparse and provides only a delayed signal, which can decrease the stability of the offline learning process. Alternatively, intermediate rewards based on medical risk scores and biomarkers provide more immediate feedback but introduce complexities (see \figureref{fig:appendix_mimic_traj_33} for an illustrative example). These noisy representations of patient health can skew the learning process, leading to policies that may not align with optimal patient outcomes, especially if the RL model disproportionately favors optimizing intermediate rewards over nuanced, long-term objectives crucial for patient care. As a result, several studies in the medical context have focused solely on rewarding RL agents based on survival outcomes, ignoring the rich set of intermediate patient health indicators collected within the ICU environment~\citep{komorowski2018artificial,peng-2018,mollura2022reinforcement}.

% Par. 3: Explain our method
To address this challenge, we propose a novel two-stage algorithm that responsibly incorporates noisy intermediate indicators into an RL framework. First, we introduce a multi-objective Q-learning algorithm that \emph{prunes} the action set based on multiple intermediate reward signals without committing to an explicit relationship between them. Second, we learn a Q-learning policy using only the sparse main reward while constraining the action space to the pruned action set derived in the first phase. In our approach, pruning relaxes the reliance on the noisy intermediate reward indicators since we focus on removing actions that are expected to be suboptimal for any linear combination of the different reward indicators. This approach minimizes the possible distortion caused by imprecise rewards and enables the learning of more effective RL policies from sparse rewards in the second stage.

% Par. 4: Off-policy evaluation
We evaluate our framework in both off-policy and offline settings using simulated environments and real data. In the off-policy setting, which includes the standard OpenAI Gym Lunar Lander environment~\citep{gym_mujoco} and the domain-specific Sepsis Simulator~\citep{oberst2019counterfactual}, our algorithm effectively prunes the action space, leading to improved policies.

% Par. 5: Offline evaluation
In an offline setting using real health records of septic patients in the ICU, our method surpasses leading offline reinforcement learning techniques Conservative Q-Learning (CQL) ~\citep{kumar2020conservative} and Discrete batch-constrained deep Q-learning (BCQ)~\citep{fujimoto2019benchmarking}. It excels both in terms of weighted importance sampling (WIS) estimates and in a qualitative assessment of the discriminative power of the Q-functions derived from various approaches. Additionally, our pruning step efficiently narrows the range of potential actions without excluding those often chosen by clinicians, indicating that the clinical relevance of the action set is maintained.

These results suggest that our two-stage algorithm is able to effectively incorporate information from diverse inputs while distinguishing between key outcomes and less accurate proxies. This capability is crucial in critical care settings, where intermediate indicators are plentiful and accurate responses to these signals are essential for providing decision recommendations. With this work, we aim to advance RL toward its potential to deliver personalized medical treatment policies tailored to the unique histories of individual patients.

The code to replicate our results is publicly available at the following link:
\begin{center}
    \url{https://github.com/alishiraliGit/multi-criteria-deep-q-learning}
\end{center}

\section{Related Work}
\label{sec:related_work}

%%%%%
\ifieee
\subsubsection*{Multi-Objective Reinforcement Learning}
\else
\paragraph{Multi-Objective Reinforcement Learning.} 
\fi
Multi-objective RL methods aim to derive policies in settings with multiple reward signals. The main body of work in this area can be divided into single-policy methods and multi-policy methods. We refer the reader to \citet{roijers2013survey} and \citet{hayes2022practical} for a survey of the field. Single-policy methods~\citep{mannor-2001,tesauro-2007} reduce multiple objectives into a single scalar reward, assuming a known user-specified or context-driven preference over different objectives, and then seek to maximize the scalarized objective using standard RL techniques. The major difference between various single-policy approaches lies in the way they attempt to derive preferences over objectives. These methods, however, cannot be used when the preferences themselves are unknown.

In contrast, multi-policy methods aim to estimate the Pareto frontier of a set of policies under different preferences. One way to achieve this is by training an ensemble of policies based on different reward scalarizations~\citep{natarajan_dynamic_2005,van-moffaert-2013,mossalam-2016,cheng2018optimal,xu-2020}. However, these methods require exhaustive training and, in some cases, non-trivial scalarizations. Another approach is a value-based method, which extends the standard scalar variables in RL algorithms to vector variables and uses updating rules in the vector space. Early work in this direction explored the problem of acquiring all Pareto optimal policies simultaneously~\citep{barrett-2008,hiraoka-2008,iima-2014}. These methods focused on applications in online settings and on small, finite-state spaces. \citet{lizotte2016multi} extended \citet{barrett-2008}'s framework to real-valued state features but with exponential complexity in the time horizon and the size of the action space. A variation of this line of work focuses on the development of non-deterministic policies (or set-valued policies) \citep{fard2011non,tang2020clinician}, which provide a set of potential solutions instead of a single action.

In this work, we build on the value-based approach to derive a pruning function using an approximate problem formulation that runs for arbitrarily long trajectories. Different from prior studies that focus on directly inferring Pareto-optimal policies, we utilize multi-objective RL to prune the action space by identifying and eliminating inferior state-action pairs. This allows us to reduce the complexity of the learning problem in the second phase of our algorithm where we train using a sparse reward alone. To our knowledge, this is one of the earliest efforts in this direction.
 
%%%%%
\ifieee
\subsubsection*{Reinforcement Learning in Health}
\else
\paragraph{Reinforcement Learning in Health.} 
\fi
Recent literature in healthcare has extensively explored RL, particularly for developing personalized treatment plans for sepsis patients in ICU settings~\citep{liu-2020, henry2015targeted, futoma-2017, raghu2017continuous, komorowski2018artificial, saria-2018, peng-2018, tang-2020, raghu2018model, mollura2022reinforcement, nanayakkara2022unifying, liu2024value, wu2023value, yu2023towards}. Many studies rely on sparse rewards \citep{komorowski2018artificial, peng-2018, mollura2022reinforcement}, with some using the intermediate SOFA score as the sole reward metric~\citep{liu2024value}, which may introduce bias in specific populations~\citep{miller2021accuracy}. Others have explored inverse RL to deduce reward functions from historical data~\citep{yu2023towards}.

In addition to the direct learning of treatment policies, another line of work has focused on detecting medical dead-end states~\citep{irpan-2018, fatemi-2019, fatemi2021medical, killian2023risk}. For instance, \citet{fatemi2021medical} developed an RL algorithm to identify and avoid states from which no action can achieve a positive terminal outcome (e.g., survival). While the aim of this method is conceptually close to the goal of our pruning stage, we propose a distinct technical approach to achieve this goal. \citet{fatemi2021medical}'s work focuses on classifying states that should be avoided; in contrast, our method directly identifies and excludes dominated \emph{state-action pairs}. Furthermore, our method does not stop at identifying risky actions but leverages the pruned action space for subsequent training of an optimal policy.

\section{Notation and Preliminaries}
\label{sec:preliminaries}

%%%%%
\ifieee
\subsubsection*{Reinforcement Learning}
\else
\paragraph{Reinforcement Learning.} 
\fi
Following the terminology commonly adopted in RL, consider an \emph{agent} interacting with an \emph{environment}. As the agent executes an \emph{action}, denoted as~$a \in \gA$, in a given \emph{state}, denoted as~$s \in \gS$, a \emph{reward}, expressed as~$r(s,a)$, materializes, and the environment's state subsequently updates to~$s'$. The agent can then utilize this reward as feedback to make better future action choices. The agent's decision-making process is described by a \emph{policy}. In general, a policy~$\pi: \gS \rightarrow \Delta(\gA)$ provides a distribution over potential actions for each state. A deterministic policy~$\pi: \gS \rightarrow \gA$ singles out a specific action given the current state. Our study particularly focuses on \emph{offline} RL, where the aim is to devise an effective policy based on previously collected data. Additionally, our algorithms are also applicable in \emph{off-policy} settings, where the agent interacts with the environment at selected time steps. We denote the available data as a set of transitions~$\gD$. Each transition is a tuple~$(s, a, s', r)$, representing that action~$a$ taken at state~$s$ results in a transition to state~$s'$ and a reward of~$r$.
%%%%%

%%%%%
\ifieee
\subsubsection*{Q-Learning}
\else
\paragraph{Q-Learning.} 
\fi
A standard (scalar) \emph{Q-function} $Q^\pi: \gS \times \gA \rightarrow \R$ estimates the value of policy~$\pi$ at state~$s$ given that the agent takes action~$a$ at~$s$ and follows policy~$\pi$ thereafter. This value represents the expected discounted sum of rewards, where rewards obtained at later timesteps are discounted by a factor of $\gamma$. If the Q-function is implemented via a deep neural network it is occasionally referred to as a \emph{Q-network}. Throughout this work, we use the terms Q-function and Q-network interchangeably. In the context of Q-Learning, we sample a batch of transitions~$\gB$ from $\gD$ and employ the Bellman equation to update the Q-network, assuming the \emph{optimal action} will be taken at subsequent steps:
\begin{equation}
\label{eq:bellman_update}
    Q(s, a) \leftarrow r + \gamma \max_{a'} Q(s', a')
    \,.
\end{equation}
Let the Q-network be parameterized by~$\theta$. Then the above notation is a shorthand to update~$\theta$ from its current value~$\theta_0$ by minimizing the loss
\begin{equation*}
    \gL(\theta)=\sum_{(s, a, s', r) \in \gB} \big(Q_\theta(s,a) - r - \gamma \max_{a'} Q_{\theta_0}(s',a')\big)^2
    \,.
\end{equation*}
This update is typically implemented via a single or a couple of gradient descent steps. However, Q-learning based on the above formulation suffers from a fast-moving target and overestimation. These challenges can be addressed by employing a target network~$Q'$ along with an update rule, recognized as double Q-learning~\citep{van2016deep}:
\begin{equation*}
    Q(s, a) \leftarrow r + \gamma \, Q'\big(s', \argmax_{a'} Q(s', a')\big)
    \,.
\end{equation*}
In double Q-learning, $Q'$ is updated by $Q' \leftarrow Q$ after hundreds or thousands of updates to~$Q$. The final deterministic policy can then be extracted as $\pi(s; Q) = \argmax_a Q(s, a)$.

%%%%%

%%%%%
\ifieee
\subsubsection*{Q-Learning With Softmax Policies}
\else
\paragraph{Q-Learning With Softmax Policies} 
\fi
In the original Bellman update, the optimal policy is deterministic, selecting the action that maximizes the Q-function for each state. However, empirical evidence has shown that applying a softmax operator tends to yield superior policies, particularly in the context of deep Q-networks. This relaxation has also garnered recent theoretical support~\citep{song2019revisiting}. Formally, given a Q-function~$Q$, we can derive a stochastic softmax policy~$\pi^\beta$ as
\begin{equation*}
    \pi^\beta(a | s; Q) = \frac{\exp\big(\beta  \, Q(s, a)\big)}{\sum_{\bar{a}} \exp\big(\beta \, Q(s, \bar{a})\big)}
    \,,
\end{equation*}
where $\beta$ is an inverse temperature parameter. Note that in the limit of~$\beta \rightarrow \infty$, the softmax policy converges to a deterministic argmax policy. Let us define
\begin{equation*}
    \softmax_{a'} Q(s', a') \coloneqq \sum_{a'} \pi^\beta(a' | s'; Q) \cdot
    Q(s', a')
    \,.
\end{equation*}
We dropped the dependence on~$\pi^\beta$ from the softmax notation for brevity. Note that $\softmax_{a'} Q(s', a') \rightarrow \max_{a'} Q(s', a')$ as $\beta \rightarrow \infty$. A Bellman update using softmax can then be obtained by substituting the $\max$ with a $\softmax$ operator in \equationref{eq:bellman_update}. Similarly, we can derive double Q-learning using a softmax policy as
\begin{equation*}
    Q(s, a) \leftarrow r + \gamma \sum_{a'} \pi^\beta(a' | s'; Q) \cdot Q'\big(s', a'\big)
    \,.
\end{equation*}
%
%%%%%

%%%%%
\ifieee
\subsubsection*{Vector-Valued Reward and Q-Function}
\else
\paragraph{Vector-Valued Reward and Q-Function.} 
\fi
In many practical scenarios, reward signals can be multi-dimensional. In particular, consider a scenario with a sparse main reward~$r_*$ and a set of $d$ noisy but more frequent auxiliary rewards. In these cases, we gather all the rewards into a \emph{vector reward}~$\vr \in \R^d$ and use~$r_i$ to refer to the $i^{\textrm{th}}$~reward. We introduce a \emph{vector-valued Q-function}~$\vQ: \gS \times \gA \rightarrow \R^d$ that outputs a vector of estimated Q-values for any state-action pair. The $i^{\textrm{th}}$~dimension of~$\vQ$, denoted by~$Q_i$, corresponds to the Q-value associated with reward~$r_i$. We discuss learning vector-valued Q-functions in \sectionref{sec:phase_1}.

%%%%%
\section{Methods}
\label{sec:methods}

\begin{figure*}[ht]
    \centering
    \includegraphics[width=0.7\textwidth]{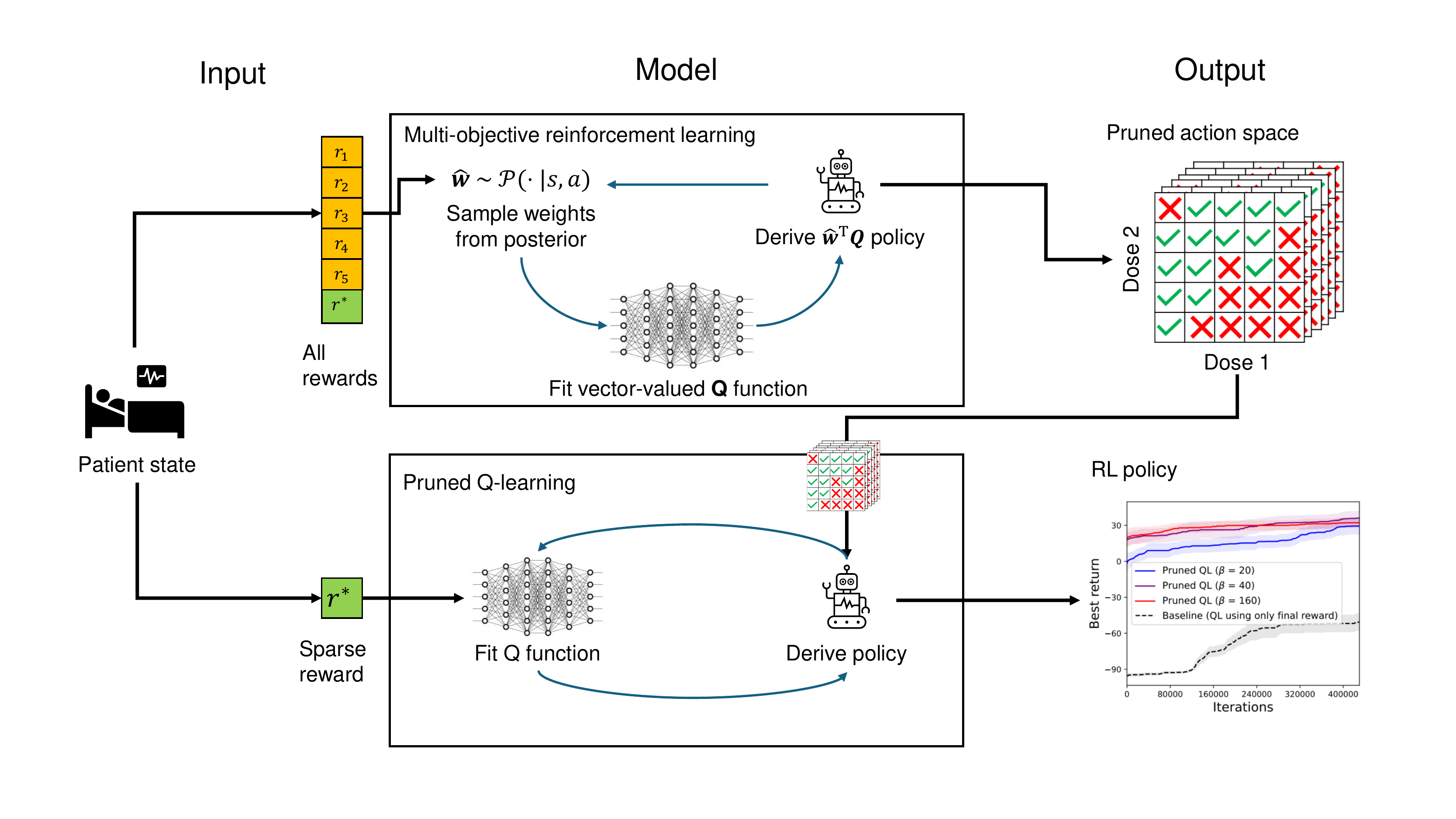}
    \caption{Illustration of our algorithm. The model first leverages all rewards in order to prune the action space for each state. Then another policy is trained based on the sparse main reward but with its action space restricted to the actions available after pruning in the first stage.}
    \label{fig:method_pictorial}
\end{figure*}

Our study aims to develop a method that effectively incorporates frequent but noisy auxiliary rewards without diverting the policy from maximizing the primary sparse reward of interest. We hypothesize that while some biomarkers or clinical risk scores might be too noisy or imprecise to be directly used in deriving an optimal policy, they may still be useful to \emph{identify actions to avoid}. Hence, we propose a two-stage algorithm: In the first phase, we learn a vector-valued Q-function relying on all rewards. This Q-function will subsequently be used to \emph{prune} the action set at each state. Then in the second phase, we search for the optimal policy based on the (accurate) sparse reward. Actions dropped in the first phase won't be available to the policy of the second phase, which reduces the complexity of the learning problem and thus facilitates learning from the sparse reward. \figureref{fig:method_pictorial} illustrates our method.

%%%%%%%%%%
\subsection{Phase 1: Multi-Objective Deep Q-Learning}
\label{sec:phase_1}
The first step of our algorithm aims to generate a pruned list of permissible actions for each state. We achieve this by leveraging several intermediate reward signals, where we use a vector-valued Q-learning method to learn optimal policies across various reward weightings. For the offline setting, we further borrow ideas from conservative Q-learning to ensure that the recommendation of the learned policy stays close to the state-action pairs observed in the dataset. The following outlines the technical implementation.

%%%%%
\ifieee
\subsubsection{A Vector-Valued Q-Learning Approach}
\else
\paragraph{A Vector-Valued Q-Learning Approach.} 
\fi
To derive a pruning policy we first need to learn a Q-function that enables us to estimate policy returns for different state-action pairs. A direct method to combine a sparse main reward and intermediate reward signals is through scalarization of the reward vector $\vr$. In its simplest form, consider a linearly weighted combination $\vw^\sT \vr$ where $\vw \in \gW \subseteq \R^d$ determines the relative importance of each reward component. We assume $w_i \ge 0$ and $\sum_{i=1}^d w_i = 1$. By using the combined reward as the standard Q-learning objective, we can obtain the Q-function $Q^\vw$ and its corresponding (possibly stochastic) policy $\pi^\beta(\cdot | \cdot; Q^\vw)$. However, the best $\vw$ may not be known a priori. To address this, we propose learning a vector-valued Q-function such that $Q^\vw(s, a) \approx \vw^\sT \vQ(s, a)$ for almost every $\vw \in \gW$. This approach avoids the need to explicitly scalarize the intermediate rewards or train a computationally expensive ensemble of models with different reward weights.

Such a Q-function cannot be learned based on a naive extension of the Bellman update. To observe this consider a fixed $\vw$ and plug $Q^\vw(s, a) = \vw^\sT \vQ(s, a)$ into the Bellman update of~$Q^\vw$:
\begin{equation}
\label{eq:mdqn_naive_update}
    \vw^\sT \vQ(s, a) \leftarrow \vw^\sT \vr + \gamma \softmax_{a'} \vw^\sT \vQ(s', a')
    \,.
\end{equation}
The above update depends on~$\vw$ and, generally, cannot hold true for all~$\vw \in \gW$: The left-hand side of the update equation is linear in $\vw$, but the right-hand side, involving a softmax operation, introduces non-linearity.\footnote{This statement also holds for deterministic policies, where the right-hand side would be piecewise linear.}

To ensure the update equation is universally applicable for any $\vw$, we approximate the softmax with a linear conservative estimate as
\begin{equation}
\label{eq:approximation}
    \softmax_{a'} \vw^\sT \vQ(s', a') \approx \sum_{a'} \pi^\beta(a' | s'; \hat{\vw}^\sT \vQ) \cdot \vw^\sT \vQ(s', a')
    \,,
\end{equation}
where $\hat{\vw}$  is sampled from a prior distribution~$\gP$. To see that this approximation is conservative, consider large values of~$\beta$. In this case, $\pi^\beta$ converges to an argmax policy, so the left-hand side of \equationref{eq:approximation} converges to $\max_{a'} \vw^\sT \vQ(s', a')$
while the right-hand side of \equationref{eq:approximation} converges to 
\begin{equation*}
     \vw^\sT \vQ(s', \argmax_{a'} \hat{\vw}^\sT \vQ(s', a'))
     \,,
\end{equation*}
serving as a lower bound for the left-hand side.

Using the approximation in~\equationref{eq:approximation}, both sides of~\equationref{eq:mdqn_naive_update} become linear in~$\vw$ and we can thus derive a vector update for~$\vQ$ that is independent of~$\vw$:
\begin{equation*}
    \vQ(s, a) \leftarrow \vr + \gamma \sum_{a'} \pi^\beta(a' | s'; \hat{\vw}^\sT \vQ) \cdot \vQ(s', a')
    \,.
\end{equation*}

\ifieee
\subsubsection{Posterior Sampling of Reward Weights}
\else
\paragraph{Posterior Sampling of Reward Weights.} 
\fi

The approximation in \equationref{eq:approximation} may be loose for a specific but unknown reward weighting $\vw$, particularly when the optimal actions for~$\vw$ and~$\hat{\vw}$ differ substantially. A domain-knowledge-informed prior distribution~$\gP$ over the space of possible weightings~$\gW$ can partially mitigate this issue by ruling out implausible weightings. We then note that for a given~$\vw$, if the probability $\pi^\beta(a | s; \vw^\sT \vQ)$ is very low, the approximation's accuracy is less critical as it underestimates the value of an action that is unlikely to be chosen. In contrast, a high probability $\pi^\beta(a | s; \vw^\sT \vQ)$ indicates the need for a precise approximation. 
This observation can be formalized by obtaining a posterior
\begin{equation*}
    \gP(\vw | s, a) \propto \gP(\vw) \cdot \pi^\beta(a | s; \vw^\sT \vQ)
    \,,
\end{equation*}
and drawing~$\hat{\vw}$ from the posterior. Intuitively, the prior is upscaled over a~$\vw$ that suggests action~$a$ is likely to be chosen at~$s$, resulting in a tighter approximation of \equationref{eq:approximation} for such weightings. Any posterior sampling method can be used to draw~$\hat{w}$. For simplicity, we use a single iteration of particle filtering, which involves sampling particles from~$\gP$, reweighting them according to $\pi^\beta$, normalizing the weights, and then resampling.

To further improve the stability and performance of the algorithm, we also use double Q-learning~\citep{van2016deep} by introducing another vector-valued Q-network~$\vQ'$ as the target network. The target network~$\vQ'$ is updated after multiple updates of~$\vQ$ by copying the weights from $\vQ$. Our update rule for $\vQ$ is
\begin{equation}
\label{eq:double_mql_update}
    \vQ(s, a) \leftarrow \vr + \gamma \sum_{a'} \pi^\beta(a' | s'; \hat{\vw}^\sT \vQ) \cdot \vQ'(s', a')
    \,,
\end{equation}
where we draw $\hat{\vw}$ from posterior~$\gP(\cdot | s, a)$.
We will refer to this method as \emph{multi-objective Q-learning (MQL)} in the following sections.
%%%%%

%%%%%
\ifieee
\subsubsection{Adaptation to Offline Learning}
\else
\paragraph{Adaptation to Offline Learning.} 
\fi
Q-learning is prone to overestimating Q-values for states and actions not present in the dataset. To overcome this challenge we borrow ideas from conservative Q-learning (CQL)~\citep{kumar2020conservative} by including an additional term in the Q-learning loss function that penalizes out-of-distribution $(s,a)$ pairs. Specifically, for a Q-function $\vQ$ parameterized by $\theta$, we calculate the loss function
\ifieee
\begin{equation*}
\begin{aligned}
    &\gL_\alpha\big(\theta; (s,a,s',\vr)\big) =  \gL\big(\theta; (s,a,s',\vr)\big) \\
    &\;\;+ \frac{\alpha}{d} \sum_{i \in [d]} \Big[ \log\big(\sum_{\bar{a} \in \gA} \exp{Q_{i,\theta}(s,\bar{a})}\big) - Q_{i,\theta}(s,a) \Big]
    \,.
\end{aligned}
\end{equation*}
\else
\begin{equation*}
\begin{aligned}
    &\gL_\alpha\big(\theta; (s,a,s',\vr)\big) =  \gL\big(\theta; (s,a,s',\vr)\big) + \frac{\alpha}{d} \sum_{i \in [d]} \Big[ \log\big(\sum_{\bar{a}  \in \gA} \exp{Q_{i,\theta}(s,\bar{a})}\big) - Q_{i,\theta}(s,a) \Big]
    \,.
\end{aligned}
\end{equation*}
\fi
Here, $\gL\big(\theta; (s,a,s',\vr)\big)$ is a standard Q-learning loss based on the update rule described in~\equationref{eq:double_mql_update}, and $i$~indexes over $d$~reward dimensions.
We will refer to this method as \emph{multi-objective conservative Q-learning (MCQL)}.
%%%%%

%%%%%
\ifieee
\subsubsection{Pruning the Action Space}
\else
\paragraph{Pruning the Action Space.} 
\fi

Using the vector-valued Q-function~$\vQ$ obtained from the MCQL or MQL method, we can now prune the action space. First, for a given~$\vQ$ and a prior~$\gP$ over reward weightings, define a stochastic policy $\pi^\beta_\gP(a | s)$ as

\begin{equation*}
    \pi^\beta_\gP(a | s) \coloneqq \E_{\vw \sim \gP}\big[ \pi^\beta(a | s; \vw^\sT \vQ) \big]
    \,.
\end{equation*}
We then propose a pruning function $\Pi^\beta: \gS \rightarrow 2^\gA$ by sampling from~$\pi^\beta_\gP$ as follows:\footnote{An alternative approach to prune the action set could be to drop actions with $\pi^\beta_\gP(a|s)$ below a threshold. However, calculating $\pi^\beta_\gP(a|s)$ requires a posterior calculation that can be computationally difficult.}

\begin{enumerate}
    \item At state~$s$, draw $m$~actions $a^{(k)} \sim \pi^\beta_\gP(\cdot | s)$ for $k \in [m]$:
    \begin{enumerate}
        \item Draw $m$~samples $\vw^{(k)} \sim \gP$ for $k \in [m]$.
        \item For each $\vw^{(k)}$, draw $a^{(k)} \sim \pi^\beta(\cdot | s; \vw^{(k)^\sT} \vQ)$.
    \end{enumerate}
    \item Set $\Pi^\beta(s) = \{a^{(k)} \mid k \in [m]\}$.
\end{enumerate}

The choice of $m$ and $\beta$ determines the expected size of the set~$\Pi^\beta(s)$. As a rule of thumb, actions with $\pi^\beta_\gP(a | s) < \frac{1}{m}$ are unlikely to remain in the action set after pruning. In practice, we set~$m$ to be at least three times the size of the action space~$|\gA|$, ensuring that all actions have a chance to appear in~$\Pi^\beta$ while still allowing rare actions to be pruned. Adjusting $\beta$ then controls the pruning strictness, with higher values leading to more deterministic policy selections. We treat~$\beta$ as a hyperparameter.

%%%%%%%%%%

%%%%%%%%%%
\subsection{Phase 2: Q-Learning with Pruning}

After obtaining a pruned action space, we employ double Q-learning with the main reward and limit actions to those selected by the pruning function $\Pi^\beta$:
\begin{equation}
\label{eq:pruned_double_q_learning_update}
    Q(s, a) \leftarrow r + \gamma \, Q'\big(s', \argmax_{a' \in \Pi^\beta(s')} Q(s', a')\big)
    \,.
\end{equation}
By using a loss function similar to conservative Q-learning~\citep{kumar2020conservative}, we can also obtain a pruned conservative Q-learning approach. 

Examining \equationref{eq:pruned_double_q_learning_update}, we observe that $\Pi^\beta(s')$ is the only place where we incorporate noisy intermediate rewards in the final policy. Optimization of the Q-function is thus solely guided by the main reward, ensuring that our learned policy remains focused on our original goal. We believe that the disentanglement of noisy and accurate rewards through this two-stage algorithm enables us to effectively incorporate information from intermediate signals to simplify the learning problem while minimally influencing the policy's primary objective.

In summary, our method termed \emph{Pruned QL} combines MQL for pruning and Q-learning for final action selection. In offline contexts, \emph{Pruned CQL} integrates MCQL and CQL.

%%%%%%%%%%
%%%%%%%%%%
\section{Off-Policy Experiments}

We begin by evaluating our method in an off-policy setting using two environments: A variation of the OpenAI Gym Lunar Lander~\citep{gym_mujoco} and the medical domain-specific Sepsis Simulator~\citep{oberst2019counterfactual}. These simulated environments allow us to directly observe the performance of the learned policy when rolled out.

Both environments are well-suited for testing our method's capabilities. In Lunar Lander, the objective is to land a small spaceship on the moon successfully. In addition to the main objective, three intermediate rewards related to the lander's shape and fuel efficiency guide the landing process. Besides the main objective, there are three intermediate rewards related to the lander's shape and fuel efficiency that guide the landing process. The combination of a sparse main objective and informative intermediate signals makes Lunar Lander an ideal choice for evaluating our approach. The evaluation results presented for this environment reflect a setting where rewards are solely based on the successful completion of the primary objective: $+100$ for a successful landing and $-100$ for failure.

The Sepsis Simulator~\citep{oberst2019counterfactual} models the simplified physiology of patients with sepsis. The actor chooses between three binary treatments: antibiotics, vasopressors, and mechanical ventilation. Each treatment can affect vital signs such as heart rate, blood pressure, oxygen concentration, and glucose levels, altering their values with predefined probabilities. Patients are discharged when all vital signs normalize and treatments cease; death occurs if three or more vitals become abnormal. Intermediate reward signals are derived from the vital sign values, while the main objective is represented by a sparse reward: +100 for patient discharge and -100 for patient death. To increase the complexity and mimic the challenge of observing final patient outcomes in real health records, we omit the final reward for a random set of $90\%$~of the trajectories. Additionally, we limit the episode length to a maximum of $20$~steps.

\begin{figure*}[ht]
    \centering
    \hspace{\fill}
    \subfigure[Lunar Lander: Best return during training]{
        \centering
        \includegraphics[width=0.38\linewidth]{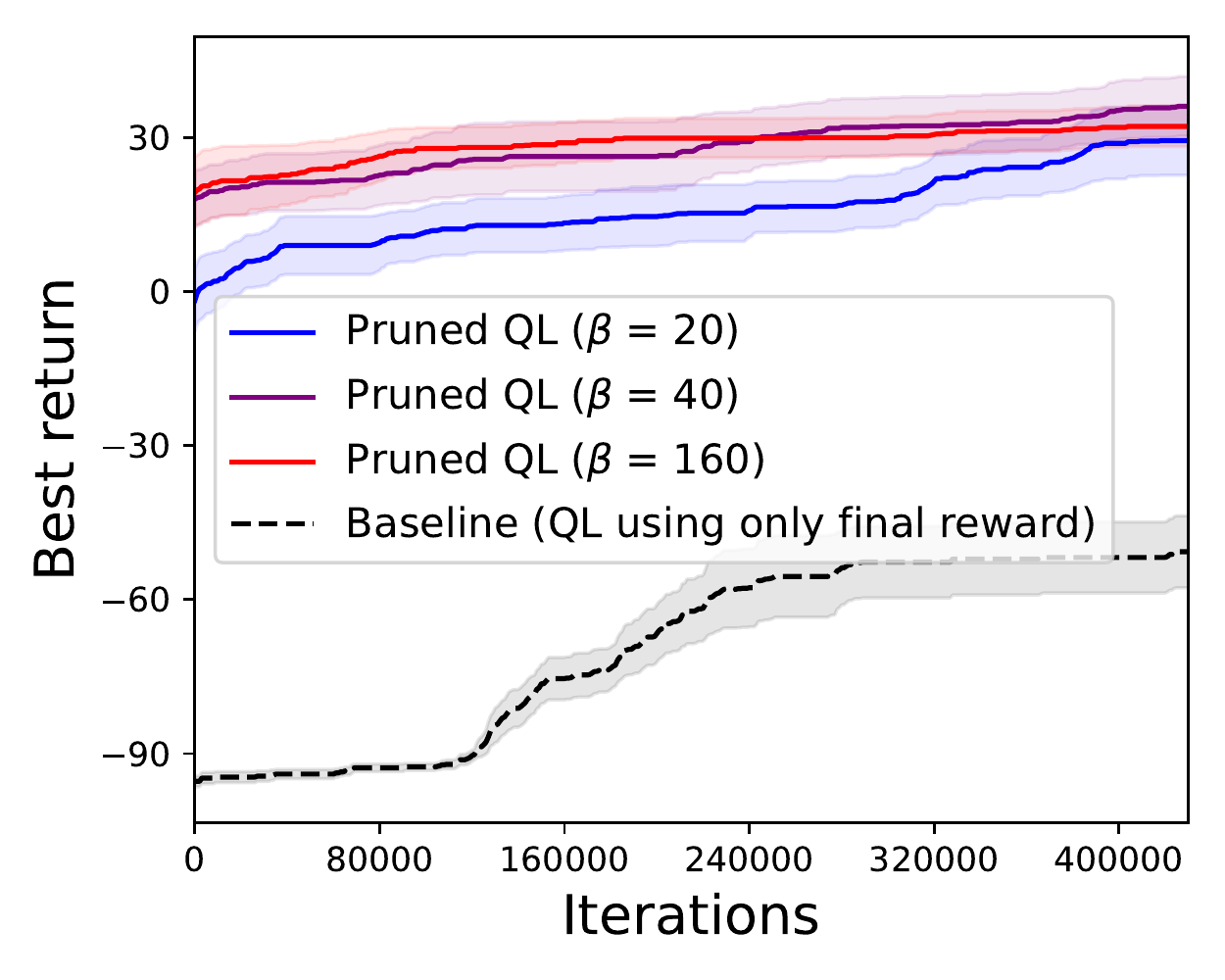}
        \label{fig:LL_train_return}
    }
    \hspace{\fill}
    \subfigure[Sepsis Simulator: Best return during training]{
        \centering
        \includegraphics[width=0.38\linewidth]{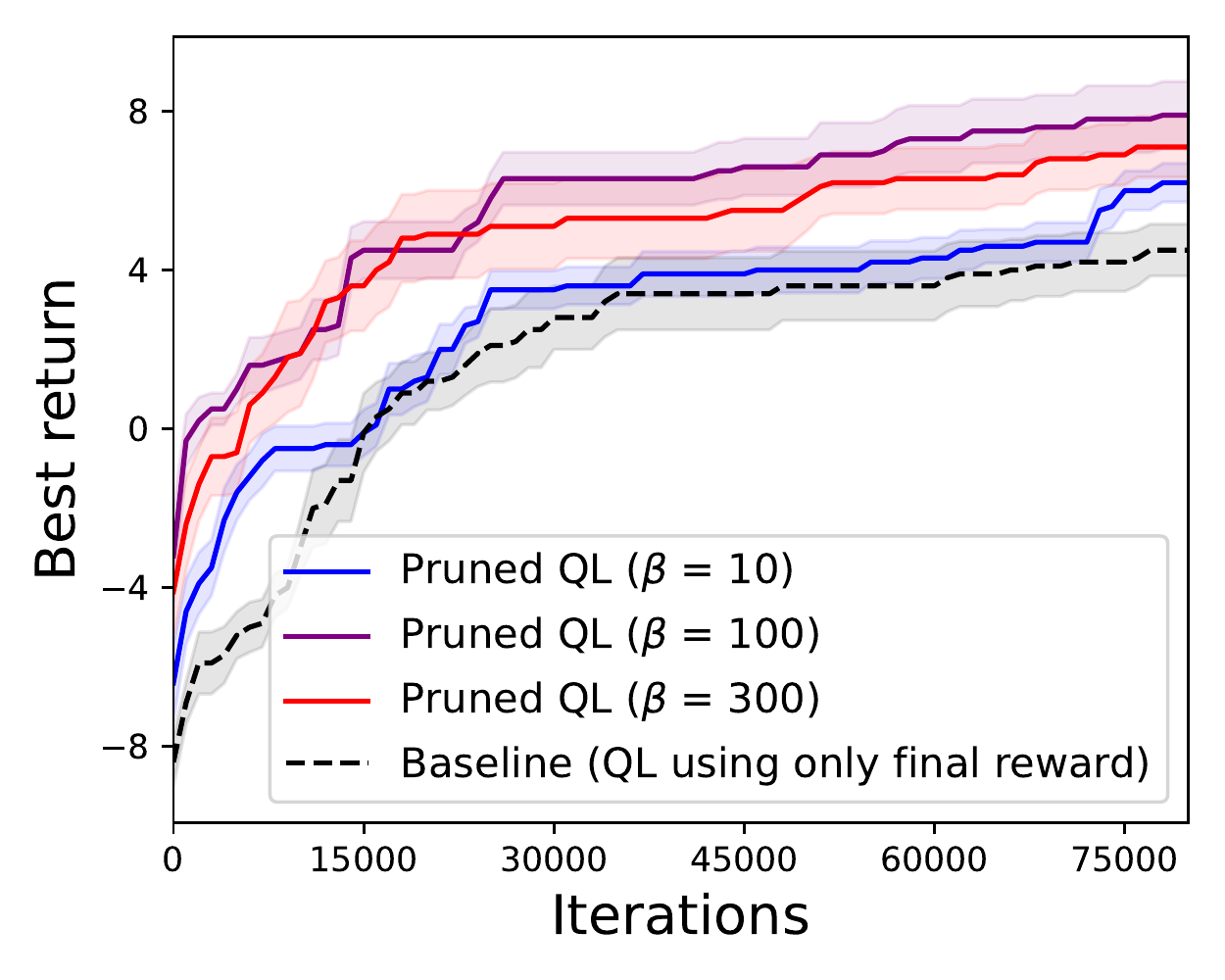}
        \label{fig:SepSim_train_return}
    }
    \hspace{\fill}
    \vskip\baselineskip
    \hspace{\fill}
    \subfigure[Lunar Lander: Policy return under different intermediate reward weightings ($\beta=40$)]{
        \centering
        \includegraphics[width=0.38\linewidth]{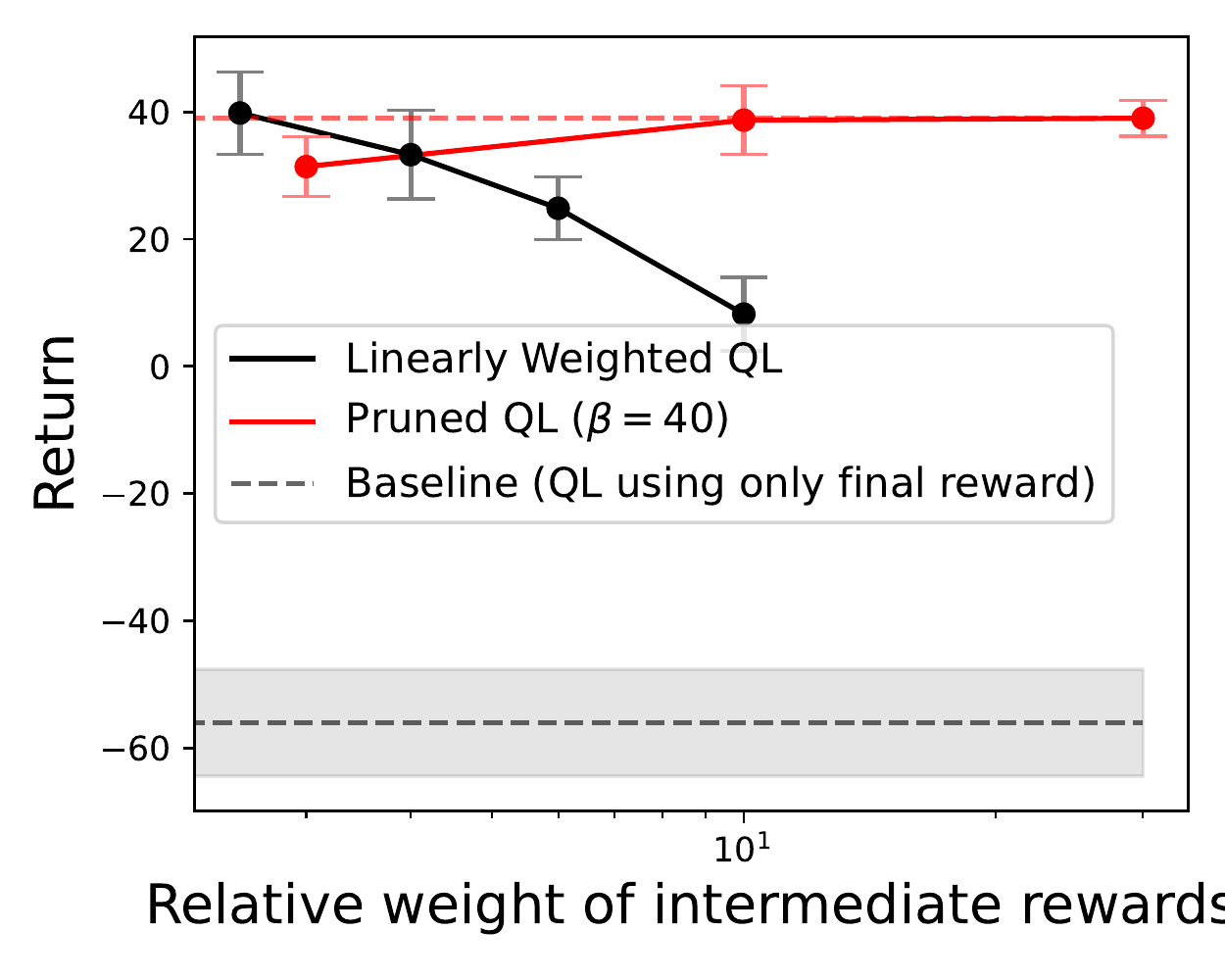}
        \label{fig:LL_weightings}
    }
    \hspace{\fill}
    \subfigure[Sepsis Simulator: Policy return under different intermediate reward weightings ($\beta=100$)]{
        \centering
        \includegraphics[width=0.38\linewidth]{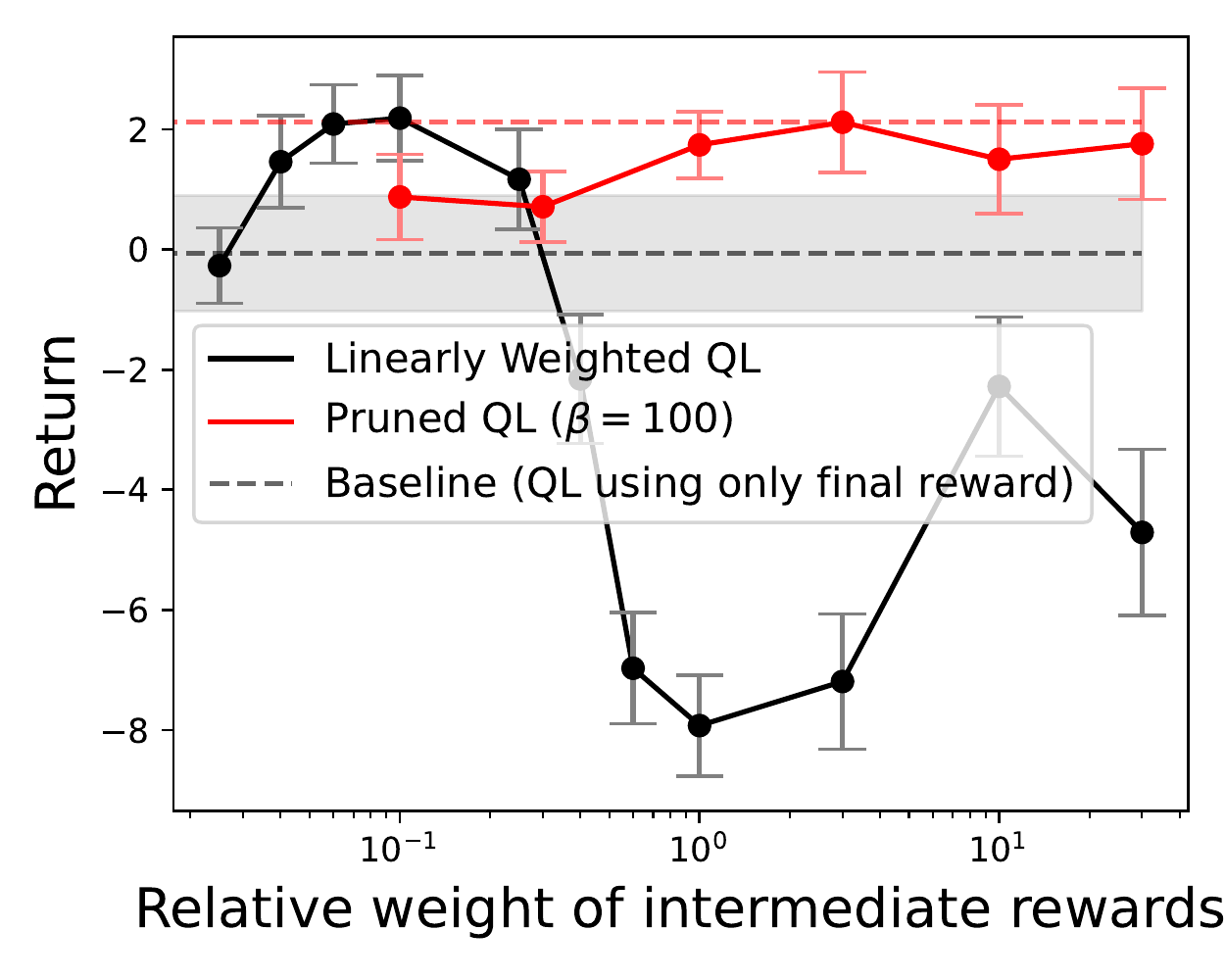}
        \label{fig:SepSim_weightings}
    }
    \hspace{\fill}
    \caption{Performance of Pruned QL in the Lunar Lander and Sepsis Simulator environments. (a) and (b) demonstrate that Pruned QL consistently achieves higher returns compared to a baseline Q-learning method that only has access to the sparse reward. Returns are estimated on new rollouts, not in the training data. (c) and (d) highlight that Pruned QL matches oracle Q-learning across varying intermediate reward weights, demonstrating its robustness and ability to leverage reward information effectively.}
    \label{fig:off_policy_results}
\end{figure*}

\figureref{fig:off_policy_results} presents the main off-policy evaluation results. We conduct two experiments for each environment. First, to analyze the performance of our method compared to a policy that only has access to the sparse reward signal, \figureref{fig:LL_train_return}~and~\ref{fig:SepSim_train_return} display the best return achieved during training for the Pruned QL algorithm under different pruning strengths, compared to a Q-learning policy trained using only sparse reward information. Our observations indicate that our method consistently outperforms the sparse reward policy across different pruning strengths in both the Lunar Lander and Sepsis Simulator environments. Notably, the Pruned QL method achieves high returns after fewer iterations and leads to policies that attain higher returns after the full 500,000 training iterations. These results demonstrate that simple Q-learning solely based on the sparse reward can be slow and ineffective in both settings. In contrast, PrunedQL effectively leverages the intermediate reward signals to derive a pruned action space using which it is possible to achieve effective learning performance from a sparse reward.

% width=0.38 for arxiv, 0.3 for ieee

We next demonstrate that Pruned QL also performs better than standard Q-learning methods when incorporating intermediate rewards during the learning process. \figureref{fig:LL_weightings}~and~\ref{fig:SepSim_weightings} plot the performance of different Pruned QL and Q-learning policies with varying strengths of intermediate reward weights. For Pruned QL, reward weights are the concentration parameters of the prior Dirichlet distribution, with higher weights indicating that intermediate rewards are more important compared to the sparse reward. By including intermediate reward data, the baseline Q-learning models now have access to the same information as the Pruned QL models.

Our results indicate that incorporating intermediate rewards does not inherently enhance model performance. Although the baseline model achieves results comparable to the Pruned QL policy when equipped with the correct (albeit typically unknown) weightings, alterations in these weightings can cause significant performance degradation. For example, in the Sepsis Simulator environment, increasing the intermediate reward weight from $0.1$ to $1$ causes the model return to decline from $2$ to $-8$, performing worse than a model trained with sparse rewards alone. In many real-world scenarios where the true reward weights are unknown, this might discourage the inclusion of intermediate reward information during model development. In contrast, the Pruned QL method demonstrates stable performance across various reward-weighting priors, effectively utilizing intermediate reward information. We further find that our method shows strong robustness to noise in the intermediate reward signals as outlined in \sectionref{sec:appendix_pql_noise}.

\ifieee
\subsubsection*{Model Parameters}
\else
\paragraph{Model Parameters.} 
\fi
We employ a Dirichlet distribution as the prior for reward weighting, with a concentration parameter of~$1$ for the main reward and $10$ for intermediate rewards. We choose a Dirichlet distribution due to its flexibility, intuitiveness, and common usage for modeling non-negative variables that sum to one, such as probabilities. As shown in \figureref{fig:SepSim_weightings}, our results are not sensitive to this parameter choice. We set the discount factor $\gamma = 1$, not discounting future rewards. The Q-function is implemented as a $3$-layer feed-forward neural network with ReLU activation, and we use the Adam optimizer with a learning rate of~$10^{-4}$. For Lunar Lander, the target network is updated once every $3000$~updates of the Q-network, while for the Sepsis Simulator, it is updated every~$10000$. For the remaining Q-learning parameters, please refer to the accompanying code.
%%%%%%%%%%

%%%%%%%%%%
\section{Offline Experiments} 

\ifieee
\subsection{Cohort and Study Design} 
\else
\subsection{Cohort and Study Design} 
\fi
We evaluate our framework in a real-life offline learning setting by training and evaluating policies for the management of vasopressor and intravenous (IV) fluids in septic patients at intensive care units (ICU). Existing works~\citep{komorowski2018artificial, peng-2018, fatemi2021medical} in this area have mainly focused on reward specifications based on $90$-day mortality. In contrast, we investigate whether Pruned CQL can facilitate learning superior policies by incorporating information from intermediate severity proxies such as the SOFA score~\citep{vincent_sofa_1996} and the patient's lactate level.

%%%%%
\ifieee
\subsubsection*{Data}
\else
\paragraph{Data.}
\fi
We use data from a cohort of septic ICU patients in the MIMIC (Medical Information Mart for Intensive Care)-III dataset (v1.4)~\citep{johnson2016mimic}. This dataset consists of deidentified electronic health records of over $40,000$~patients who were admitted to the critical care units of the Beth Israel Deaconess Medical Center in Boston between 2001 and 2012. To construct our sample, we followed the preprocessing steps applied in \citet{komorowski2018artificial}. We excluded patients younger than $18$~years old at ICU admission or where mortality and intravenous fluid intake have not been documented. A trajectory includes data from $24$~hours before the sepsis onset up until $48$~hours after the onset, recorded in $4$-hour intervals. For every trajectory, we extracted a set of $48$~variables, including demographics, Elixhauser comorbidity index~\citep{elixhauser1998comorbidity}, vital signs, laboratory test results, and medication dosing decisions. All models are trained on $80\%$ of the data, validated on $5\%$, and tested on $15\%$ for the final evaluation.

%%%%%

%%%%%
\ifieee
\subsubsection*{Action Set and State Space}
\else
\paragraph{Action Set and State Space.}
\fi
Following previous works~\citep{komorowski2018artificial,fatemi2021medical}, we discretize actions into $25$~treatment choices ($5$~discrete levels for IV fluids and $5$~discrete levels for vasopressor dosage). To derive the state space, we applied K-means clustering to the $48$-dimensional features and obtained $752$~clusters. 
As the resulting cluster indicators were not numerically meaningful, we additionally applied the continuous bag of words method~\citep{mikolov2013efficient} to the patients' state trajectories. This provided us with a $13$-dimensional representation for each cluster. Thus, our problem formulation is based on a $25$-dimensional action space and $752$ states, where each state is represented by a $13$-dimensional vector. The resulting dataset consists of $20,912$ ICU stay trajectories, of which $4,917$ resulted in patient death within $90$~days of the patient's critical care visit.

%%%%%

%%%%%
\ifieee
\subsubsection*{Reward Specification}
\else
\paragraph{Reward Specification.}
\fi
We assign a reward of~$+100$ to the final state of a trajectory if the patient survives for at least $90$~days past ICU admission, and a reward of~$-100$ otherwise. Achieving a low $90$-day mortality is the ultimate goal of our learning agent but since this is a sparse and delayed signal, we further include four medically-motivated rewards that are observed more frequently throughout the patient's stay:

\begin{itemize}
    \item \textbf{One-period SOFA score change}: The negative of the one-period change in SOFA score.
    \item \textbf{Two-period SOFA score change}: The negative of the two-period change in SOFA score.
    \item \textbf{One-period lactate level change}: The negative of the one-period change in the lactate level.
    \item \textbf{Two-period lactate level change}: The negative of the two-period change in the lactate level.
\end{itemize}
SOFA~\citep{vincent_sofa_1996} is a medical risk score that summarizes the extent of a patient’s organ failure and in recent years has become a key indicator of the sepsis syndrome~\citep{lambden2019sofa}. Arterial lactate levels are an important biomarker for septic shock because they are closely associated with cellular hypoxia. Sepsis can cause an imbalance in oxygen supply and demand, resulting in inadequate delivery of oxygen to the cells and tissues. This can lead to anaerobic metabolism and the production of lactate as a byproduct. Increased arterial lactate levels, therefore, indicate that there is a mismatch between oxygen supply and demand and that the body is experiencing hypoxia~\citep{gernardin1996blood}.  

%%%%%

%%%%%
\ifieee
\subsubsection*{Model Parameters}
\else
\paragraph{Model Parameters.}
\fi
The model hyperparameters in the offline setting are largely analogous to those in the off-policy setting. Additionally, we select a CQL-alpha parameter of $0.001$. The target Q network is updated every $1000$ training steps during the first phase and every $8000$ steps during the second phase. For further implementation details, please refer to the code repository associated with this paper.
%%%%%

%%%%%
\ifieee
\subsection{Policy Evaluation Approach}
\else
\subsection{Policy Evaluation Approach}
\fi

In the offline RL setting, rolling out the learned policies to observe their returns is not feasible. Therefore, we evaluate the policies using weighted importance sampling (WIS) and a descriptive measure to assess the Q-functions' ability to distinguish between high and low mortality trajectories.

Various methods exist for estimating policy values in offline settings, each with limitations. For an empirical comparison of different methods, see \citet{tang2021model}, and for practical considerations, see \citet{gottesman2018evaluating}.

Importance sampling methods are an effective class of techniques for policy evaluation in the offline setting. The main idea behind these evaluation approaches is to re-weight trajectories based on their likelihood of occurrence to estimate the policy value. This requires a stochastic policy and knowledge of the behavioral policy. To accommodate our deterministic policy in phase~$2$, we follow \citet{tang-2020} by \emph{softening} our policy: $\tilde{\pi}(a | s) = (1 - \epsilon) \One\{a = \pi(s)\} + \frac{\epsilon}{|\gA| - 1} \One\{a \neq \pi(s)\}$, where $\epsilon$ is set to~$0.01$. To calculate the importance ratio, we approximate the physician's policy with a stochastic policy using a multi-class logistic regression with cross-entropy loss. We train this classifier to predict the next action based on the the $13$-dimensional state space. We then implement WIS to estimate policy values due to its lower variance compared to ordinary importance sampling.

One drawback of the WIS estimator is its potential bias. This bias arises because the estimator conservatively relies on the observed data generated by the behavior policy, which may not adequately represent the target policy's distribution. Additionally, normalizing or clipping the importance weights to control variance further introduces bias, skewing the estimates towards trajectories that are more common under the behavior policy.

To supplement the WIS evaluation, we propose a descriptive measure that evaluates our Q-function's ability to differentiate between patient survival and death trajectories. This is relevant since a high capacity of a Q-function to distinguish between low and high-risk state-action pairs increases our confidence that choosing the actions with the highest predicted Q-value is beneficial. We adopt the following approach to construct such a descriptive measure. Consider a trajectory~$\tau$ from the test data and let $r_*^\tau$ be the mortality-based reward assigned to its final state. We compute $Q(s, a)$ for each state-action pair $(s, a) \in \tau$. Note that $Q(s, a)$ is supposed to be the expected return from the optimal policy at state~$s$ followed by action~$a$ and not the return of the physician's policy~$r_*^\tau$. But if the optimal policy is sufficiently similar to the physicians, $Q(s, a)$ shall reflect $r_*^\tau$, and we would like to see a larger $Q(s, a)$ if $\tau$ results in the patient survival. To measure this association, we bin the Q-values into quartiles and calculate the mortality rate in the lower quartile range $MR(Q_1)$ and the mortality rate in the upper quartile range $MR(Q_3)$. We then report the difference $\Delta MR = MR(Q_1) - MR(Q_3)$. A larger $\Delta MR$ indicates that the policy is more effective in distinguishing survival from death trajectories.
%%%%%

%%%%%
\ifieee
\subsection{Offline Evaluation Results}
\else
\subsection{Offline Evaluation Results}
\fi

We evaluate Pruned CQL against standard conservative Q-learning (CQL)~\cite{kumar2020conservative} and discrete batch-constrained Q-learning (BCQ)~\cite{fujimoto2019benchmarking}. Discrete BCQ also uses a threshold rule to constrain the action space of its policy, allowing only actions whose probability relative to the most likely action under the behavior policy exceeds a certain threshold $t$. However, it focuses on enforcing similarity to the behavior policy rather than leveraging intermediate reward signals as our method. Additionally, Discrete BCQ relies on an explicit estimation of the behavior policy.

\tableref{tab:delta_mr_wis_and_physician_accuracy} summarizes the offline evaluation results, demonstrating Pruned CQL's superior performance in terms of both $\Delta MR$ and WIS. The Pruned CQL policy with $\beta = 160$ achieves a WIS of~$66$, surpassing the behavior policy (WIS of $51$) and nearly doubling the best CQL policy's performance (WIS of $35$). Additionally, Pruned CQL consistently outperforms Discrete BCQ in both WIS and $\Delta MR$. Discrete BCQ's $\Delta MR$ is about $10$~percentage points lower than CQL or Pruned CQL, suggesting a potentially inflated WIS value due to its similarity with the behavior policy. Additional findings in appendix \sectionref{sec:appendix_mimic_q_func} illustrate that stratifying trajectories based on the Q-values predicted by Pruned CQL allows to effectively distinguish between death and survival trajectories.

\begin{table*}[h]
    \centering
    \caption{Comparison of different methods in terms of~$\Delta MR$, WIS-based policy value, and the share of common actions between the learned and the behavioral policy. Standard errors are calculated based on ten random seeds. Best result per column is underlined. The behavior policy has a return of~$51.9$.}
    \begin{tabular}{cccc}
        \toprule
        Model & $\Delta MR$ (\%) & Value (WIS) & Behavior policy overlap (\%)  \\
        \midrule
        CQL($\alpha=0.001$) & $24.6 \pm 1.0$ & $14 \pm 18$ & $10.4 \pm 0.7$ \\
        CQL($\alpha=0.005$) & $23.6 \pm 1.0$ & $35 \pm 21$ & $18.6 \pm 0.6$ \\
        CQL($\alpha=0.01$)  & $22.6 \pm 0.9$ & $26 \pm 25$ & $26.2 \pm 1.0$ \\
        \midrule
        Discrete BCQ ($t=0.05$) & $12.9 \pm 0.8$ & $31 \pm 26$ & $36.5 \pm 0.3$ \\
        Discrete BCQ ($t=0.1$) & $13.8 \pm 0.7$ & $47 \pm 24$ & $39.6 \pm 0.4$ \\
        Discrete BCQ ($t=0.3$)  & $14.8 \pm 1.1$ & $51 \pm 22$ & \underline{$42.8 \pm 0.3$} \\
        \midrule
        Pruned CQL($\alpha=0.001$, $\beta=20$)  & $24.5 \pm 0.9$ & $32 \pm 12$ & $11.3 \pm 1.2$ \\
        Pruned CQL($\alpha=0.001$, $\beta=40$)  & \underline{$25.2 \pm 0.6$} & $41 \pm 15$ & $15.6 \pm 0.9$ \\
        Pruned CQL($\alpha=0.001$, $\beta=160$) & $24.2 \pm 1.1$ & \underline{$66 \pm 19$} & $22.1 \pm 0.9$ \\
        \bottomrule
    \end{tabular}
    \label{tab:delta_mr_wis_and_physician_accuracy}
\end{table*}

\begin{figure}[ht]
    \centering
    \hspace{\fill}
    \subfigure{
        \label{fig:delta_mr_vs_physician_accuracy}
        \includegraphics[width=0.38\linewidth]{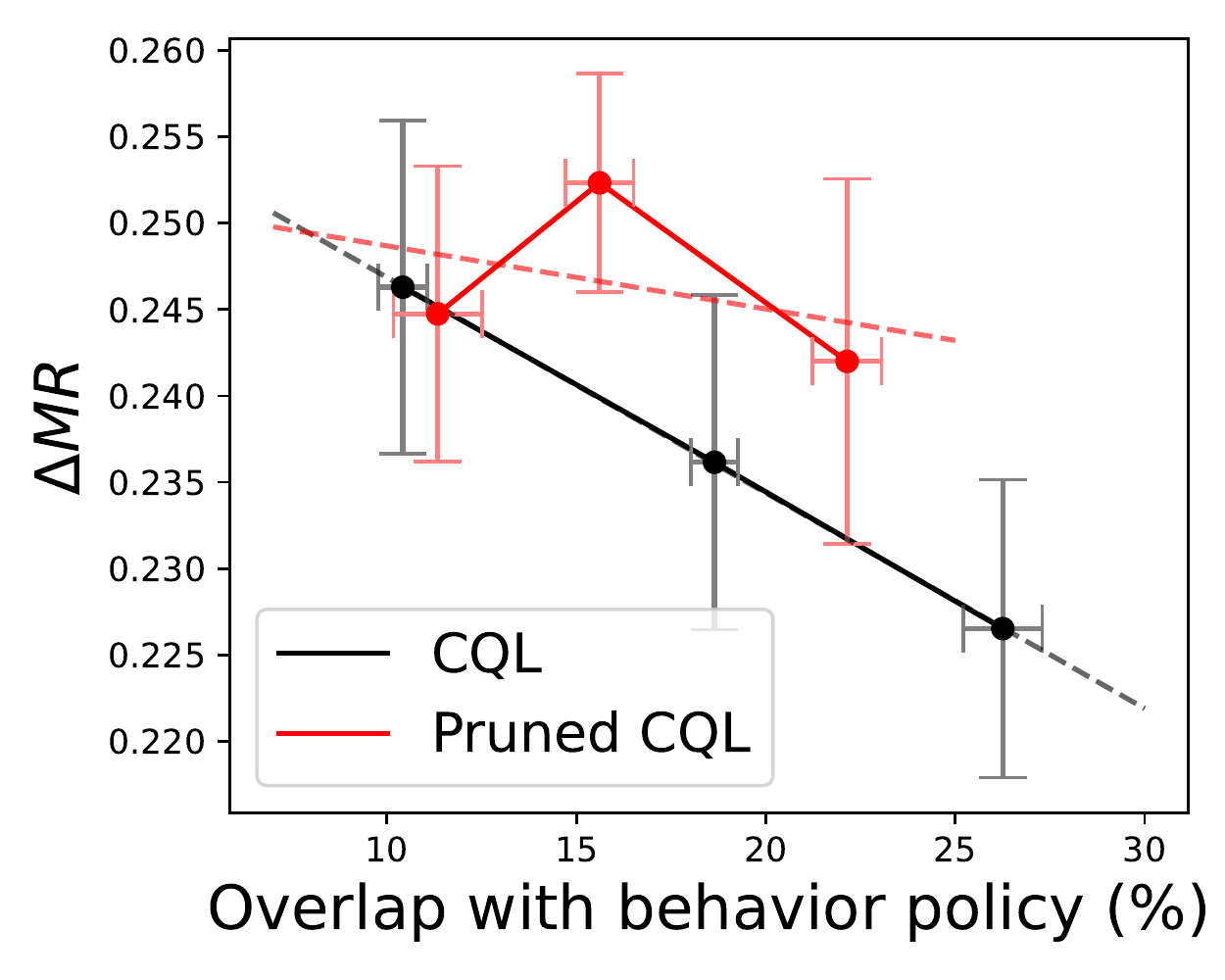}
    }
    \hspace{\fill}
    \subfigure{
        \label{fig:wis_vs_physician_accuracy}
        \includegraphics[width=0.38\linewidth]{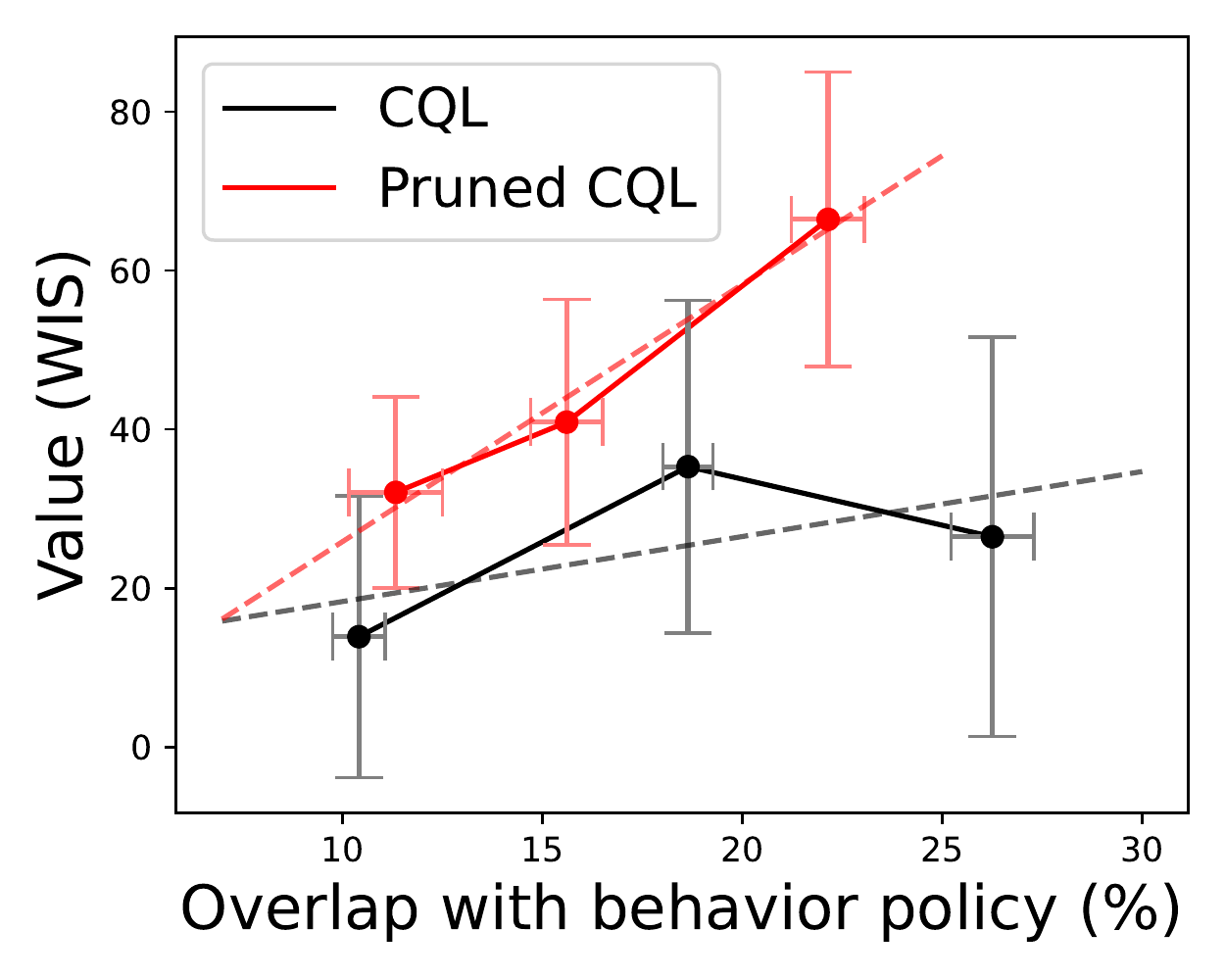}
    }
    \hspace{\fill}
    \caption{Comparison of Pruned CQL and CQL in terms of $\Delta MR$ and WIS-based policy value, for different degrees of overlap with the behavior policy. Dashed lines display linear fits.}
    \label{fig:delta_mr_wis_vs_physician_accuracy}
\end{figure}

\tableref{tab:delta_mr_wis_and_physician_accuracy} demonstrates that stricter pruning leads to policies that closely align with physician behavior. This alignment is likely driven by the pruning step enforcing greater consideration of key intermediate severity indicators crucial for physician decision-making. Similarly, higher conservativity~$\alpha$ in Baseline CQL also yields greater agreement. \figureref{fig:delta_mr_wis_vs_physician_accuracy} illustrates this relationship for varying CQL-$\alpha$ and pruning strength $\beta$. We generally observe that increasing these hyperparameters enhances performance in terms of WIS, while maintaining or reducing performance in $\Delta MR$, likely due to WIS's bias towards the behavior policy. Furthermore, Pruned CQL consistently outperforms baseline CQL in both $\Delta MR$ and WIS-value at similar agreement levels. These findings suggest that our pruning approach is an effective alternative to the standard CQL loss, to increase safety by closely aligning the learned RL policy with physician practices.

% width=0.38 for arxiv, 0.45 for ieee

%%%%%

%%%%%
\ifieee
\subsection{Pruning Analysis}
\else
\subsection{Pruning Analysis}
\fi

To evaluate the quality of our pruning method, we address two key questions: 1) Does our pruning procedure in phase~$1$ significantly reduce the size of the action space? 2) Is the resulting action space consistent with current medical practice?

\begin{table}[b]
    \centering
    \caption{Mean action set size and recall for different pruning levels. The initial action set size was $25$.}
    \begin{tabular}{ccc}
        \toprule
        $\beta$ & \makecell{Num. of available actions\\after pruning} & Recall (\%)  \\
        \midrule
        $20$  & $19.7\pm 0.3$  & $94.7 \pm 0.3$ \\
        $40$  & $11.6 \pm 0.7$ & $83.3 \pm 1.4$ \\
        $160$ & $4.1 \pm 0.3$  & $49.4 \pm 1.9$ \\
        \bottomrule
    \end{tabular}
    \label{tab:pruned_size_recall}
\end{table}

\tableref{tab:pruned_size_recall} presents the average number of available actions after pruning and the corresponding recall of these action sets. Since there is no ground truth for the best policy, we define recall based on whether the pruned action sets contain the actions taken by physicians in each respective state in the test set. Given that physicians are highly trained professionals and may have access to additional indicators not captured in our records, our goal is to retain their decisions among the available options as much as possible. Our results indicate that the pruning procedure significantly reduces the size of the original $25$-dimensional action space: Moderate pruning with $\beta=40$ reduces the mean action set size to less than half, while stricter pruning with $\beta=160$ reduces the average number of available actions to nearly one-sixth. Despite such aggressive pruning, our method maintains a high recall, ranging from $49\%$ to over $94\%$, well beyond the chance level.

\begin{figure}[h!]
    \centering
    \includegraphics[width=0.6\linewidth]{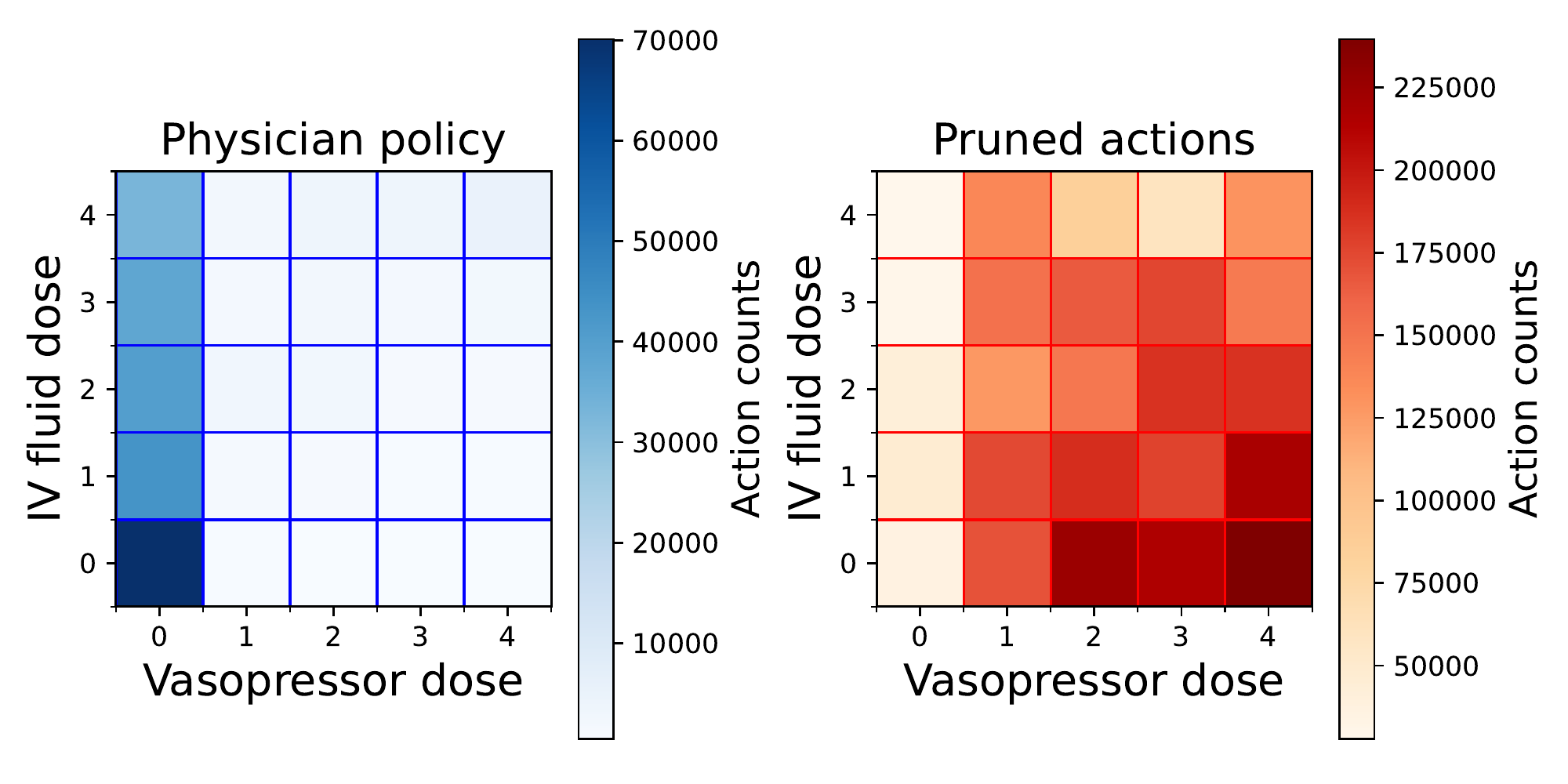}
    \caption{Distribution of physician's actions and pruned actions ($\beta$=40).}
    \label{fig:action_dist_alpha_40}
\end{figure}

To further investigate the pruning behavior, \figureref{fig:action_dist_alpha_40} shows the distribution of removed actions compared to the actions taken by physicians in the test set for a pruning strength of $\beta=40$. Similar patterns are observed for both less stringent and more stringent pruning levels, as detailed in \figureref{fig:pruned_actions_distribution_for_various_betas}. The figure indicates that our procedure primarily prunes extreme dosing regimes or incoherent decisions, such as administering a low intravenous fluid dose while simultaneously assigning a high vasopressor dose. These findings support the notion that the pruning procedure reduces the complexity of the reinforcement learning problem while retaining relevant actions for the subsequent learning stage.

% width=0.6 for arxiv, 0.9 for ieee

\section{Conclusion and Discussion}

In this study, we introduced a novel RL method that utilizes action space pruning to facilitate learning when the main reward signal is sparse, intermediate signals are approximate, and there is ambiguity about the optimal reward component weights. Our algorithms effectively integrate more frequent but imprecise reward proxies into learning.

We demonstrated the effectiveness of our approach in off-policy settings using the Lunar Lander and Sepsis Simulator environments, where action-space pruning enables Q-learning from sparse rewards alone, outperforming standard Q-learning approaches. Furthermore, in an offline setting prescribing vasopressor and intravenous fluid dosing for septic ICU patients, our method learned more effective policies compared to leading offline RL methods, including Conservative Q-learning and Discrete BCQ. Our learning framework efficiently incorporates recurrent yet imprecise reward signals while preventing such signals from causing the policy to diverge from maximizing the primary reward of interest. Although motivated by healthcare applications, our approach is applicable to other domains where the true reward signal is sparse and available proxies are frequent but imperfect indicators of the outcome of interest.

\ifieee
\subsubsection*{Limitations}
\else
\paragraph{Limitations.} 
\fi
First, our results are based on data up to 2012, and sepsis treatment guidelines have evolved since then, meaning the estimated physician policies may reflect a slightly outdated standard of care. Second, our method relies on the availability of meaningful intermediate reward signals that can serve as proxies for the final outcome of interest, and it may not be applicable in cases where such signals are not available. Finally, we utilized weighted importance sampling (WIS) for offline evaluation. Although WIS is a widely accepted technique for evaluating RL policies in offline settings, it is also known to have high variance and is dependent on accurate estimation of the behavior policy~\citep{tang2021model}. We partially addressed this issue by evaluating our policy using a related descriptive measure that highlights consistent results with the provided WIS estimates.

\pagebreak

\printbibliography

\ifieee
\else
    \clearpage
    \appendix
\fi

%%%%%%%%%%
\section{Different Biomarkers Along Patient Trajectory}
\label{sec:appendix_pt_traj_rewards}

\begin{figure}[h]
    \centering
    \includegraphics[width=0.5\linewidth]{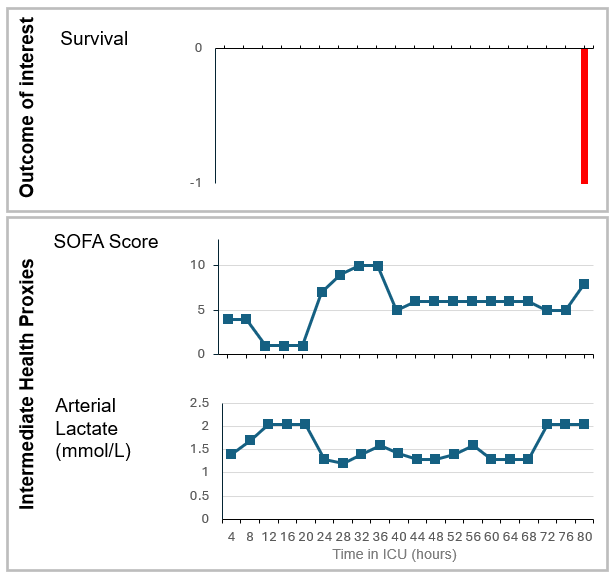}
    \caption{Intermediate and sparse reward signals along the patient trajectory.}
    \label{fig:appendix_mimic_traj_33}
\end{figure}

To demonstrate the advantages and limitations of the intermediate reward signals, \figureref{fig:appendix_mimic_traj_33} presents the survival, SOFA score, and arterial lactate data gathered across a specific patient trajectory. The SOFA score and arterial lactate values offer a detailed view of the patient's health progression. However, they are not perfectly correlated with mortality; thus, optimizing these parameters alone may not suffice for learning the optimal policy.

% width=0.5 for arxiv, 0.7 for ieee
%%%%%%%%%%

%%%%%%%%%%%
\newpage
\section{Pruned Q-Learning Under Different Noise Levels}
\label{sec:appendix_pql_noise}

\begin{figure}[h]
    \centering
    \hspace{\fill}
    \subfigure[$\beta = 20$]{
        \label{fig:lunar_lander_learning_curves_vary_noise_20}
        \includegraphics[width=0.38\linewidth]{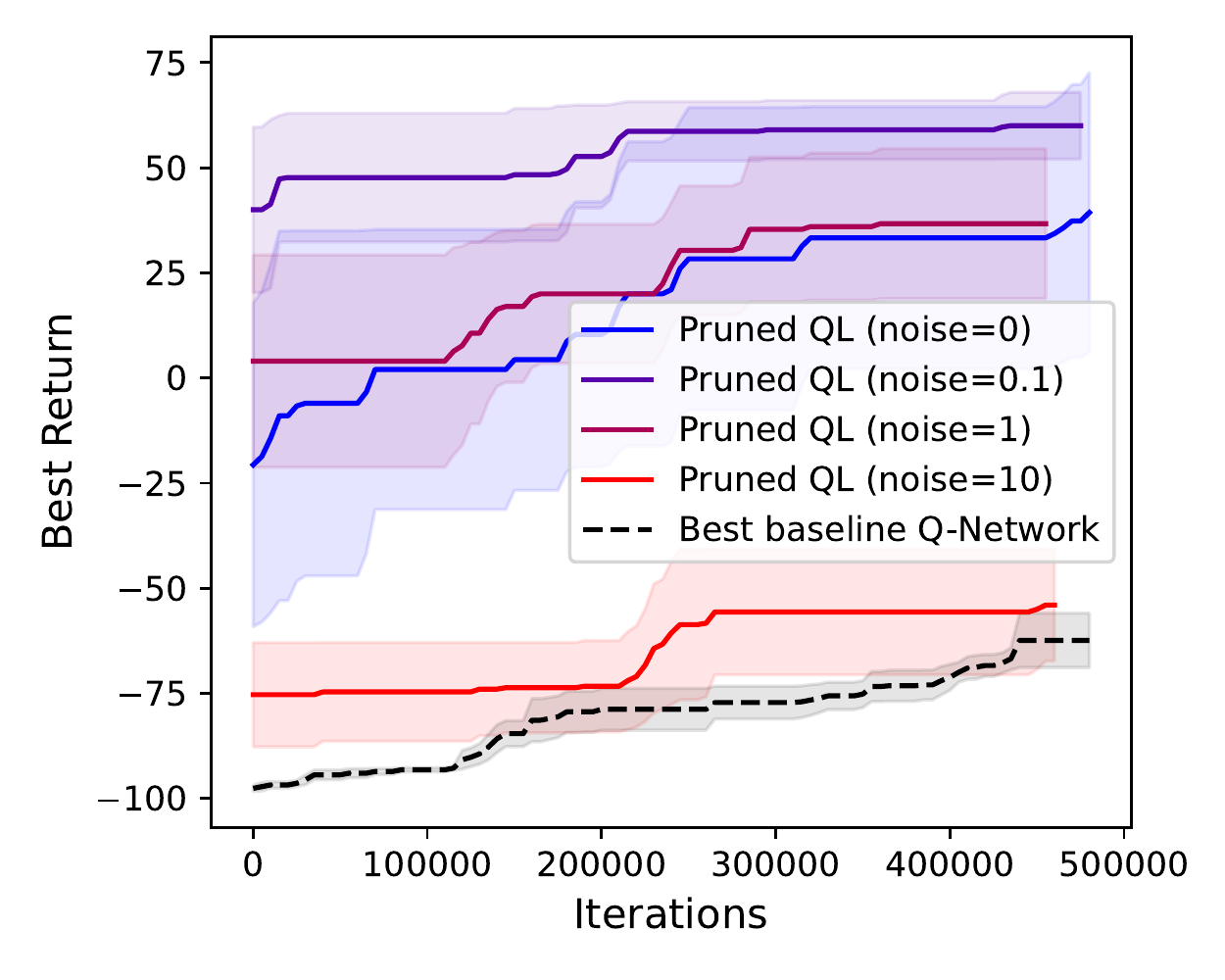}
    }
    \hspace{\fill}
    \subfigure[$\beta = 40$]{
        \label{fig:lunar_lander_learning_curves_vary_noise_40}
        \includegraphics[width=0.38\linewidth]{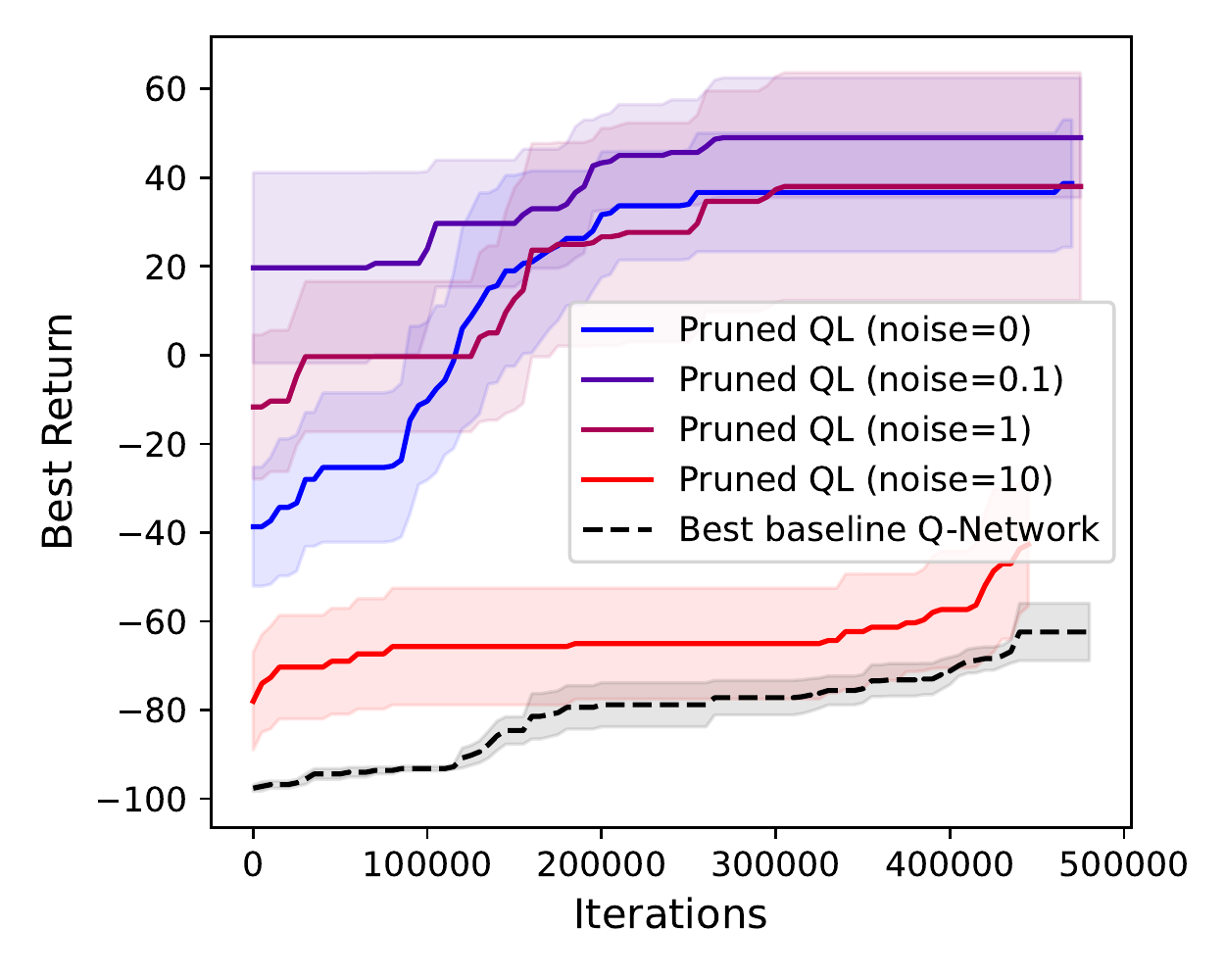}
    }
    \hspace{\fill}
    \vskip\baselineskip
    \subfigure[$\beta = 160$]{
        \label{fig:lunar_lander_learning_curves_vary_noise_160}
        \includegraphics[width=0.38\linewidth]{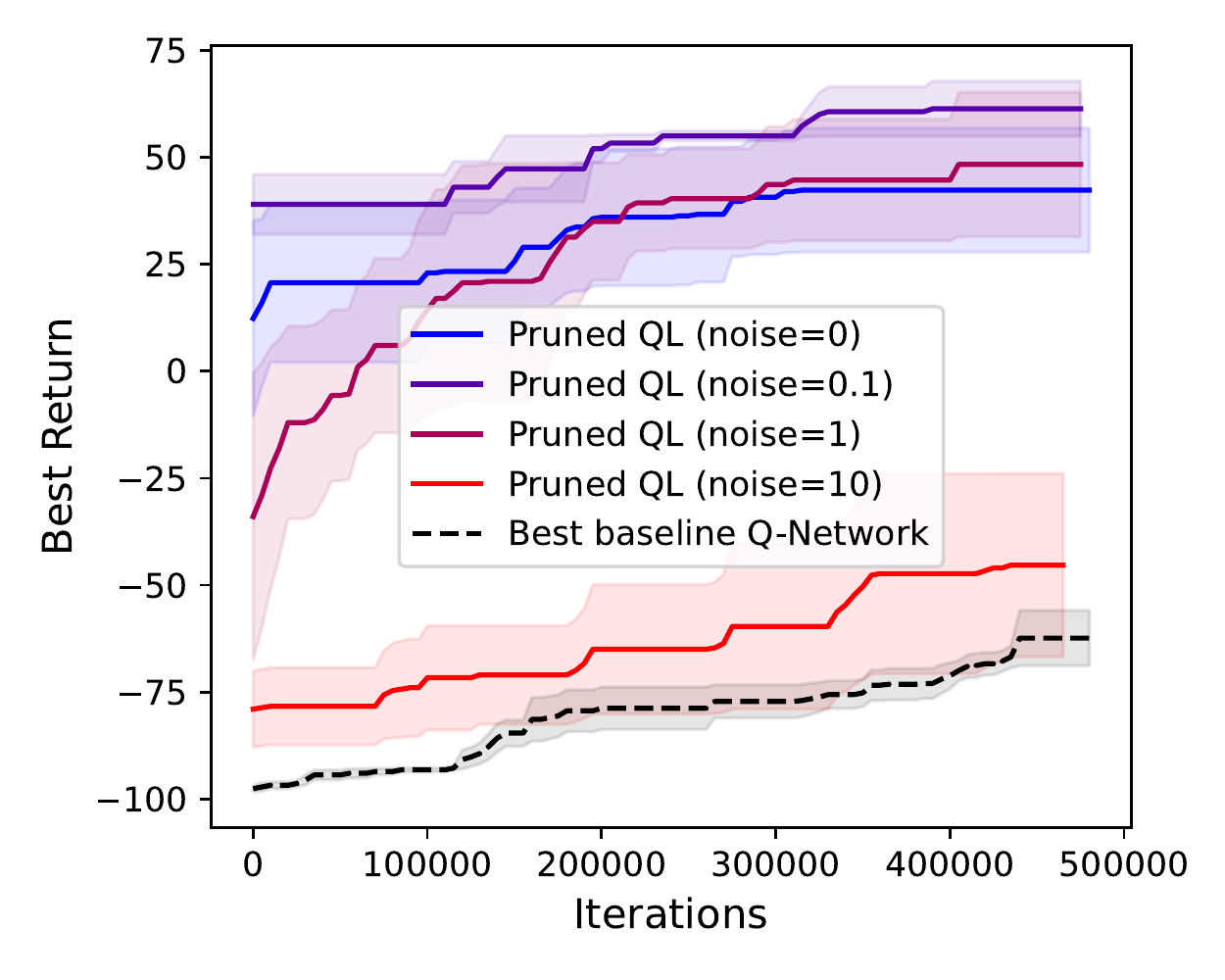}
    }
    \caption{Learning curves indicating policy performance of Pruned QL vs. best baseline Q-Network under different noise levels.}
    \label{fig:lunar_lander_experiments_noise_various_beta}
\end{figure}

We analyze how Pruned QL behaves as the intermediate reward signal becomes less precise. To achieve this we conduct an experiment where we add white noise with standard deviations of $0.1$, $1$, and $10$ to the intermediate rewards. \figureref{fig:lunar_lander_experiments_noise_various_beta} depicts the results. We observe that Pruned QL remains robust until raising the white noise standard deviation to $10$, where it only does marginally better than the baseline. This result illustrates two important characteristics of our method. First, it shows that Pruned QL has generally high robustness to noise. Second, the fact that the performance drops when the noise standard deviation is increased to~$10$ indicates that the intermediate rewards considered in this method must carry some relevant information about the sparse outcome of interest. At a white noise standard deviation of~$10$, the signal-to-noise ratio (SNR) of the shape reward is less than~$0.5$ and the SNR for the fuel rewards is less than~$0.02$, diluting nearly the entirety of the reward signal. This result is robust for various pruning strengths.

%%%%%%%%%%

%%%%%%%%%%
\newpage
\section{Further Insights From MIMIC-III Experiments}

%%%%%
\subsection{Survival Rate by Q-value for different pruning strengths}
\label{sec:appendix_mimic_q_func}

\begin{figure}[h]
    \centering
    \hspace{\fill}
    \subfigure[$\beta=20$]{
        \includegraphics[width=0.38\linewidth]{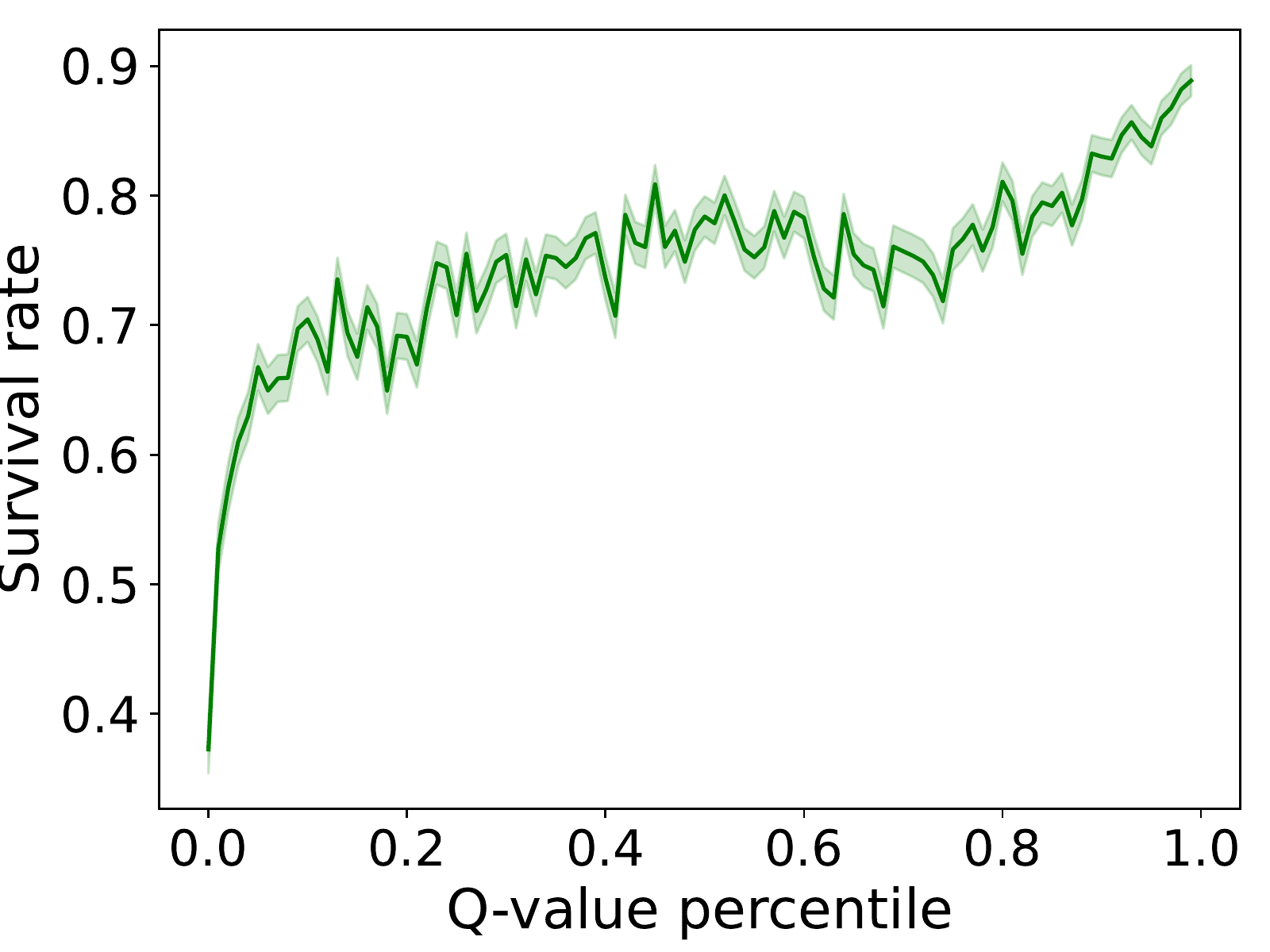}
        \label{fig:survival_Q_alpha20}
    }
    \hspace{\fill}
    \subfigure[$\beta=40$]{
        \includegraphics[width=0.38\linewidth]{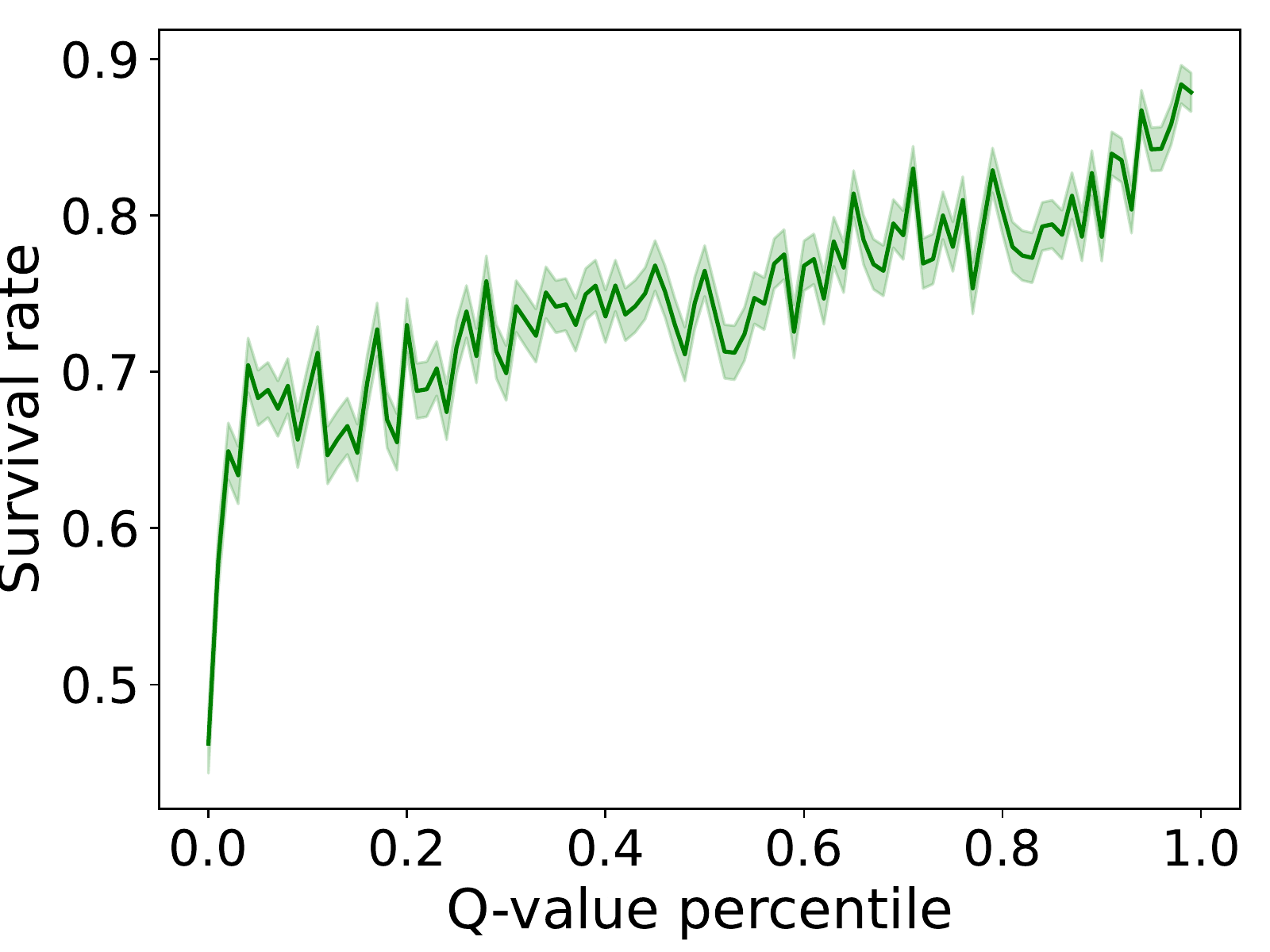}
        \label{fig:survival_Q_alpha40}
    }
    \hspace{\fill}
    \vskip\baselineskip
    \subfigure[$\beta=160$]{
        \includegraphics[width=0.38\linewidth]{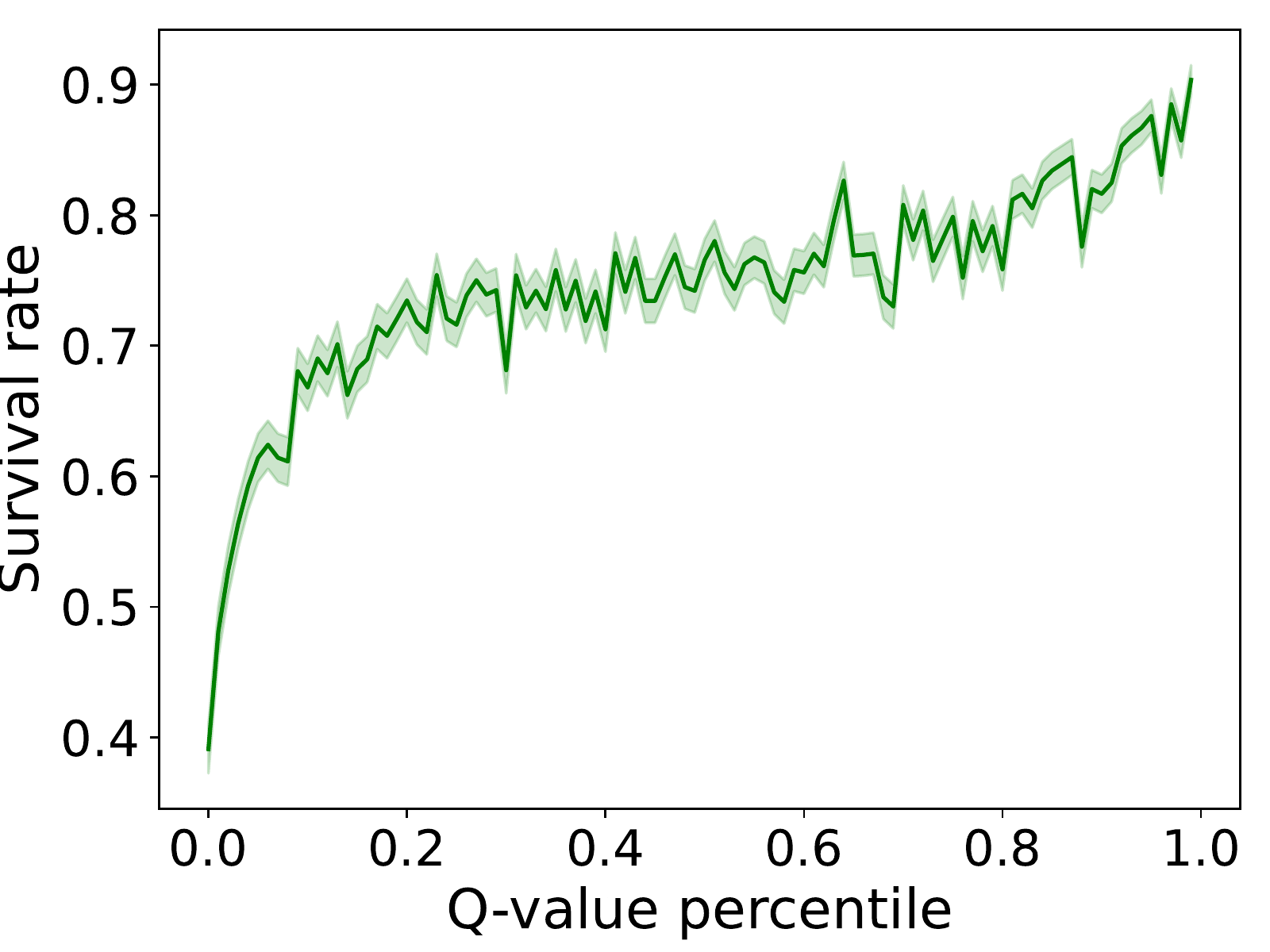}
        \label{fig:survival_Q_alpha160}
    }
    \caption{Survival rate by Pruned CQL Q-value percentile plotted for $100$ equal-sized bins and different pruning strength. }
    \label{fig:combined_survival_Q}
\end{figure}

To further analyze the Q-values obtained from Pruned CQL, we calculated the mean Q-values for each patient trajectory in the test data and divided them into 100 percentile bins. For each bin, we determined the probability of patient survival. \figureref{fig:combined_survival_Q} presents the survival rate per bin for Pruned CQL with a conservativity level of $\alpha=0.001$ and varying pruning strictness. The analysis demonstrates a strong positive relationship across different pruning strengths. For instance, with a pruning strength of 40, we observe a survival rate below 30\% for state-action pairs with the lowest Q-values and over 80\% for those with the highest Q-values. These results indicate that the Pruned CQL policies are based on Q-functions that effectively distinguish between high- and low-risk state-action pairs, which is crucial for identifying relevant policies.

%%%%%

%%%%%
\newpage
\subsection{Pruned Action Distribution for Different Pruning Strengths}
\label{sec:appendix_mimic_exp_pruning}

\begin{figure}[h]
    \centering
    \subfigure[$\beta=20$]{
        \label{fig:action_dist_alpha5}
        \includegraphics[width=0.6\linewidth]{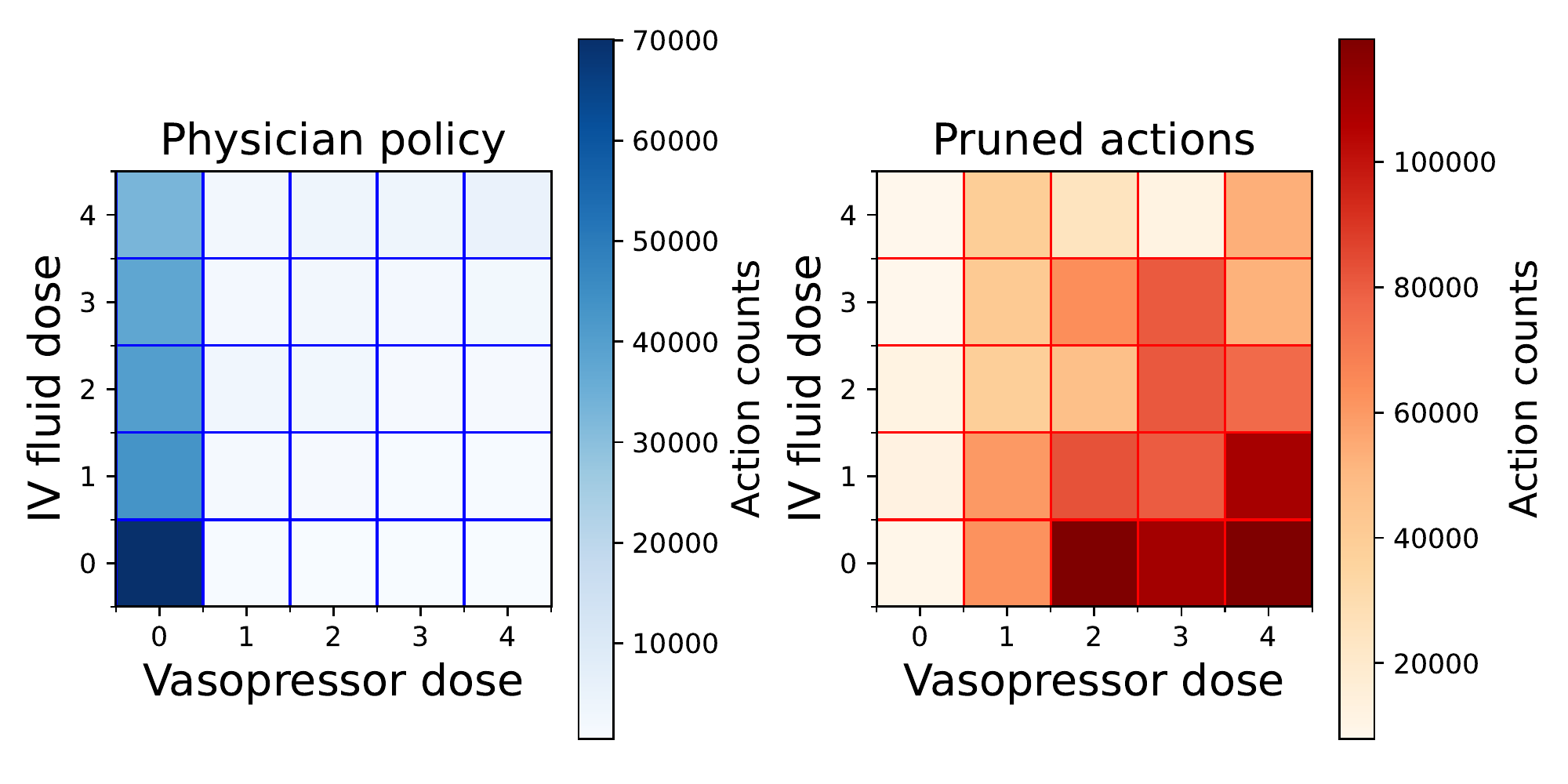}
    }
    \hfill
    \subfigure[$\beta=160$]{
        \label{fig:action_dist_alpha20}
        \includegraphics[width=0.6\linewidth]{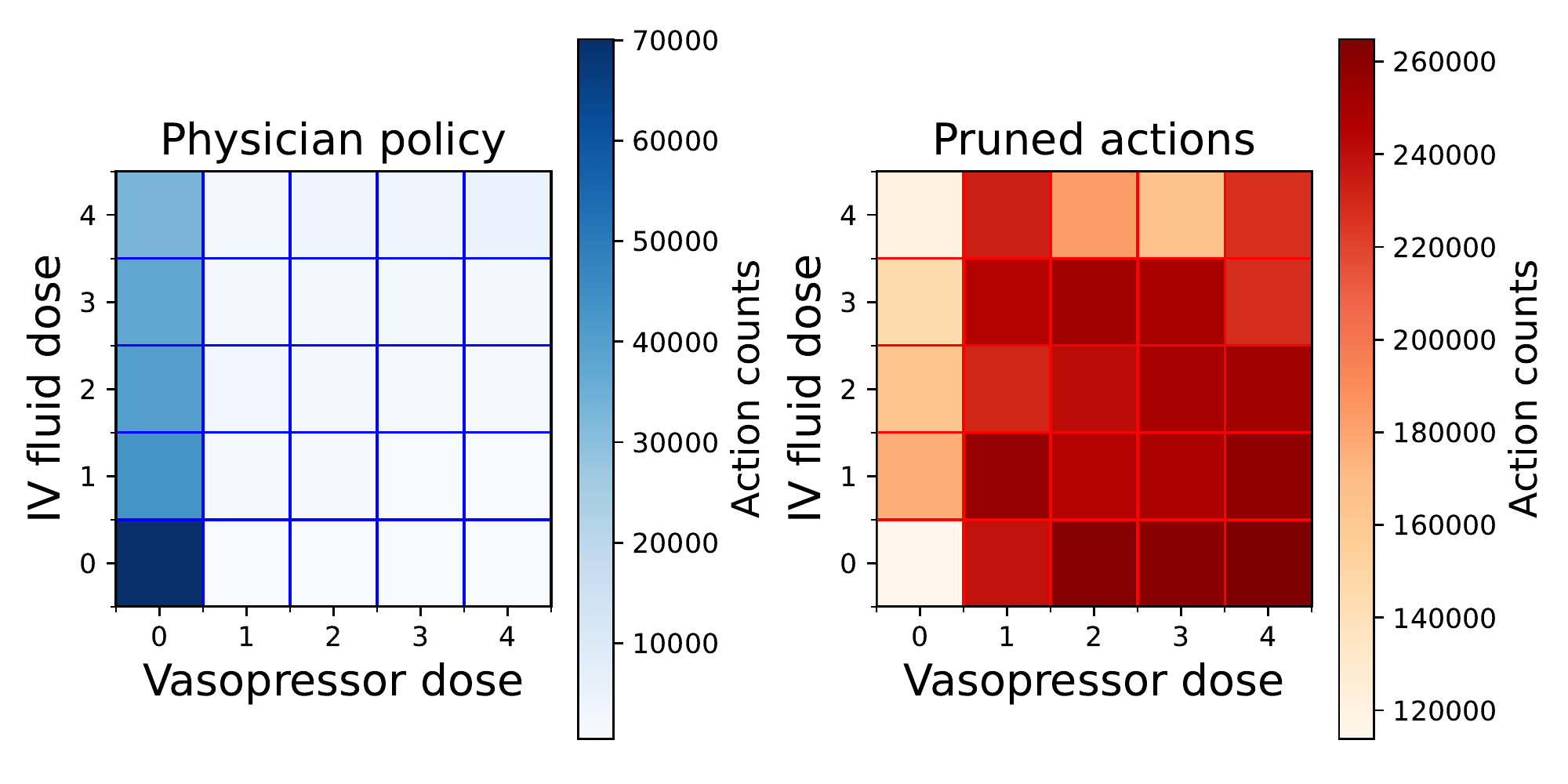}
    }
    \caption{Distribution of pruned actions for different pruning strengths.}
    \label{fig:pruned_actions_distribution_for_various_betas}
\end{figure}

To augment the findings illustrated in \figureref{fig:action_dist_alpha_40}, we depict the distribution of removed actions for the pruning strength values $\beta=20, 160$ in \figureref{fig:pruned_actions_distribution_for_various_betas}.

% width=0.6 for arxiv, 0.7 for ieee

%%%%%

\end{document}